\documentclass[runningheads]{llncs}

\usepackage{eccv}

\usepackage{eccvabbrv}

\usepackage{graphicx}
\usepackage{booktabs}

\usepackage[accsupp]{axessibility}  

\newcommand{\Tref}[1]{Table~\ref{#1}}
\newcommand{\eref}[1]{Eq.~(\ref{#1})}

\newcommand{\fref}[1]{Fig.~\ref{#1}}

\newcommand{\sref}[1]{Sec.~\ref{#1}}
\newcommand{\Sref}[1]{Section~\ref{#1}}

\newcommand{\textblock}[1]{\noindent\textbf{#1}}

\def\eg{\emph{e.g}\onedot}

\usepackage{multirow}
\usepackage{enumitem}
\usepackage{makecell}
\usepackage{amsmath}
\usepackage{amssymb}
\usepackage{graphicx}
\usepackage{subcaption}
\usepackage{gensymb}
\usepackage{gensymb}
\usepackage{adjustbox}
\usepackage{wrapfig}

\usepackage{hyperref}

\usepackage{orcidlink}

\begin{document}

\title{Hierarchically Structured Neural Bones \\ 
for Reconstructing Animatable Objects
\\ from Casual Videos} 

\titlerunning{Hierarchically Structured Neural Bones}

\author{Subin Jeon\orcidlink{0000-0003-1651-2249} \and
In Cho\orcidlink{0009-0006-2131-4430} \and
Minsu Kim\orcidlink{0009-0004-3359-0198} \and
Woong Oh Cho\orcidlink{0009-0002-5059-7258} \and
Seon Joo Kim\orcidlink{0000-0001-8512-216X}}

\authorrunning{S.~Jeon et al.}

\institute{Yonsei University
}

\maketitle
\begin{abstract}
We propose a new framework for creating and easily manipulating 3D models of arbitrary objects using casually captured videos.
Our core ingredient is a novel hierarchy deformation model, which captures motions of objects with a tree-structured bones. Our hierarchy system decomposes motions based on the granularity and reveals the correlations between parts without exploiting any prior structural knowledge.
We further propose to regularize the bones to be positioned at the basis of motions, centers of parts, sufficiently covering related surfaces of the part.
This is achieved by our bone occupancy function, which identifies whether a given 3D point is placed within the bone.
Coupling the proposed components, our framework offers several clear advantages: (1) users can obtain animatable 3D models of the arbitrary objects in improved quality from their casual videos, (2) users can manipulate 3D models in an intuitive manner with minimal costs, and (3) users can interactively add or delete control points as necessary.
The experimental results demonstrate the efficacy of our framework on diverse instances, in reconstruction quality, interpretability and easier manipulation.
Our code is available at \url{https://github.com/subin6/HSNB}.
\end{abstract}
\begin{figure*}[t]
    \centering
    \includegraphics[trim={0 20 0 0pt}, width=1\linewidth]{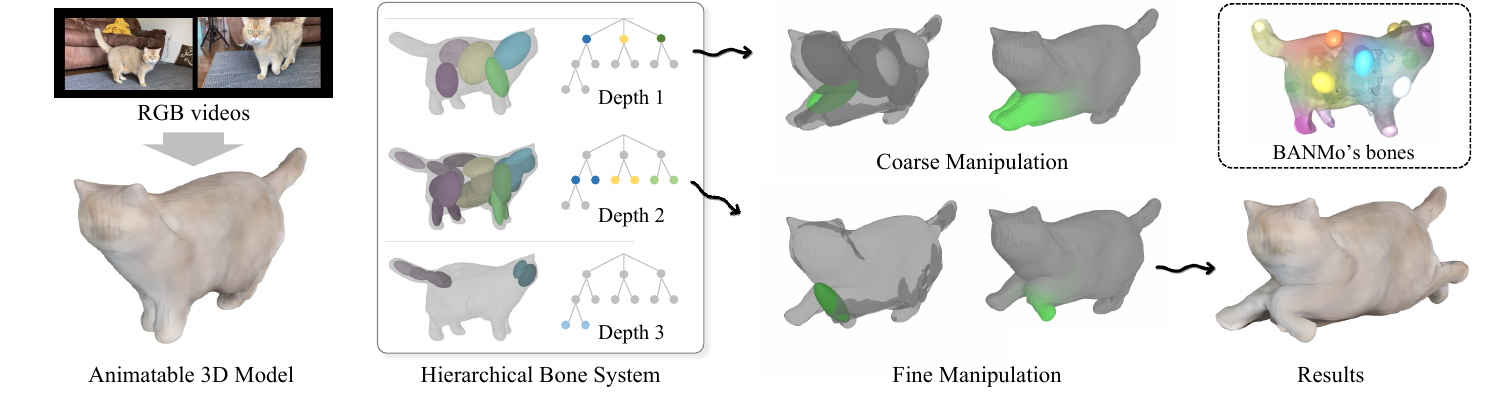}
    \captionof{figure}{
    We aim to reconstruct animatable models that can be manipulated in a coarse-to-fine manner, using multiple videos capturing a deformable object. 
    The resulting 3D model can be manipulated using a hierarchical deformation model, where coarse motions are manipulated using the parent bones, and fine motions are subdivided by the child bones.
    We present manipulation results in novel poses.
    }
    \label{fig:intro}
\end{figure*}

\section{Introduction}
We have witnessed rapid development in creating animatable 3D models, which are playing vital roles in diverse industries, \eg films, mixed reality, and games.
However, such development is primarily carried out at the industry level, requiring enormous labor costs and a level of proficiency.
Most of the general users, on the other hand, remain distant from this industry-level advancement, demanding more simplified ways to obtain animatable models.
Recent methods \cite{yang2022banmo, yang2023rac, peng2021animatable-nerf, li2022tava} have suggested an alternative yet effective approach for general users: building animatable models from casually captured videos.

These methods employed the framework of Neural Radiance Fields (NeRF) \cite{mildenhall2021nerf} with various forms of controllable deformation models to handle the motions between frames.
While a number of research \cite{peng2021animatable-nerf, li2022tava, peng2021neuralbody, weng2022humannerf} adopted predefined or hand-crafted templates, \eg skeletons \cite{li2022tava, peng2021animatable-nerf, yang2023rac} and 3D body models \cite{weng2022humannerf, peng2021neuralbody}, we stand for utilizing a set of Gaussian ellipsoids as control points, as in BANMo \cite{yang2022banmo}.
These ellipsoids, so-called bones, offer a way to acquire articulated 3D models without being constrained to prior knowledge.
Despite their general applicability, utilizing these bones as ``control points'' poses challenges in actual manipulation.
This is due to the absence of structures, as these bones are distributed across the object surfaces without considering the granularity of movements, lacking correlations between bones with similar motions.
Such an unstructured property also leaves room for improvement in reconstruction quality, often requiring plenty of input videos to produce plausible results.

In this paper, we present a framework for creating and easily manipulating 3D models of arbitrary objects from casual videos.
We build our framework upon BANMo \cite{yang2022banmo}, with careful consideration to tailoring control points into well-structured forms.
To provide better understanding of motions and facilitate easier manipulation of the reconstructed objects, our structured deformation model aims to decompose the motions, capture shared movements based on the granularity, and identify correlations among parts with similar motions.

To achieve this goal, we introduce a novel hierarchical bone system that represents object deformations with tree-structured bones.
Our key idea is to learn the deformations in a coarse-to-fine manner: parent bones capture coarse motions of broader regions, with each child bone representing finer motion at a more specific part.
We begin with a small number of bones, covering coarse parts, and gradually append child bones to cover finer motions of more specific parts.
The resulting tree-structured bones identify connections between relevant bones in a fully unsupervised manner.
These connections facilitate users to easily understand the structures of the motions and provide better interpretability, as well as improving reconstruction quality.

Furthermore, we suggest a regularization approach where bones are positioned at the centers of their respective parts. This is achieved using bone masks derived from the bone occupancy function and foreground masks of the objects.
Instead of placing them around the surfaces as in previous methods, we extend the concept widely used in part-based generative methods \cite{genova2020lif, tertikas2023partnerf, paschalidou2021neuralparts} into our reconstruction pipeline for animatable models.
Our bone regularization term prevents surfaces of the same part from being assigned to different bones.
This facilitates our hierarchical bones to correctly capture the parts sharing motions, ensuring each bone can serve as a basis for the motions of each part.

Coupling these key ideas, the structured control points in our framework provide a more user-friendly tool for creating and manipulating 3D models with several clear advantages:
\begin{itemize}
    \item Obtaining animatable 3D models of improved quality from casual videos.
    \item Manipulating 3D models in an intuitive manner with minimal effort.
    \item Interactively adding or deleting control points in desired parts.
\end{itemize}
We evaluate the effectiveness of our method through extensive experiments on various instances, showcasing high-quality results of the models as well as interpretable and structured control points.
We also demonstrate the manipulation capability of our framework through reanimation and manipulation results.

\section{Related Work}
\textblock{Dynamic 3D Reconstruction. }
Dynamic reconstruction \cite{dou2016fusion4d, collet2015high, bozic2020deepdeform, innmann2016volumedeform, zollhofer2014realtime, lin2022occlusionfusion} aims to reconstruct per-frame 3D geometry from a given video sequence.
Recently, inspired by NeRF \cite{mildenhall2021nerf}, its dynamic variant have significantly improved this field using only RGB videos.
These dynamic methods, known as Dynamic NeRFs, can be broadly categorized into two streams.
Firstly, deformation-based methods \cite{tretschk2021nr-nerf, pumarola2021d-nerf, park2021nerfies, park2021hypernerf} learn canonical NeRFs and per-frame deformation fields from the observation space to the canonical space simultaneously.
Another line of approaches \cite{xian2021spacetime, li2021nsff, gao2021dynamic, li2022dynerf, cao2023hexplane, fridovich2023kplanes} involve learning time-conditioned NeRFs, which take time and 3D position as input and directly output color and density.
Despite impressive results of such dynamic methods, the implicit learning of deformations makes it challenging to manipulate scenes into novel poses.

\textblock{Animatable Object Reconstruction. }
Reconstructing animatable objects is a longstanding challenge in computer vision and graphics.
Its goal is to reconstruct 3D models with accurate geometry that can be manipulated into novel poses.
Category-specific approaches have been extensively studied with category-level templates.
Model-based methods \cite{alldieck2019peopleinclothing, alldieck2018humanavatars, alldieck20183dpeople, peng2021neuralbody, liu2021neuralactor} represent input motions using 3D deformable models \cite{loper2023smpl, anguelov2005scape, xiang2019monocular, pavlakos2019expressive, blanz2023morphable}, while skeleton-based methods \cite{su2021a-nerf, li2022tava, yang2023rac, peng2021animatable-nerf, weng2022humannerf, wu2022casa} utilize skeletons.
Recent advances in NeRF have also spurred active research in these approaches \cite{liu2021neuralactor, peng2021neuralbody, su2021a-nerf, li2022tava, yang2023rac, peng2021animatable-nerf, weng2022humannerf}.
However, acquiring these category-specific templates necessitates either extensive 3D scan data or thorough annotation of the respective category.
Such templates limits general applicability of these methods across diverse types of objects.

On the other hand, category-agnostic methods \cite{yang2021lasr, yang2021viser, yang2022banmo, kuai2023camm} propose reconstructing animatable 3D objects from videos by learning control points simultaneously with the 3D shape, bypassing the need for predefined templates. 
Among these methods, BANMo \cite{yang2022banmo} demonstrates promising results using a NeRF-based 3D model and linear blend skinning with implicitly learned bones. 
Successive researches have attempted to improve BANMo in various aspects, including root pose decomposition \cite{wang2023root}, incomplete view coverage \cite{tu2023dreamo}, and text-to-4D generation \cite{wang2023animatabledreamer}.
Orthogonal to these attempts, our method aims to improve deformation modeling to address challenge of manipulation capability, which is a crucial aspect but has received relatively little attention.

\textblock{Part-level representation.}
Contrary to research that utilizes low-level primitives to represent motions of objects, there have been studies \cite{genova2019sif, genova2020lif, paschalidou2021neuralparts, hertz2022spaghetti, tertikas2023partnerf, noguchi2022watch} focusing on learning partial representations of 3D shapes using low-level primitives such as ellipsoids, spheres, and cubes.
In these approaches, objects are composed of multiple primitives, where each primitive represents semantic parts of the object and models shapes of it.
The part-based generative models are learned from a collection of data on the single class, aiming for shape abstraction \cite{genova2020lif}, part understanding \cite{paschalidou2021neuralparts, genova2019sif}, and part-based shape editing \cite{hertz2022spaghetti, tertikas2023partnerf}.
We draw inspiration from this line of works to regularize our deformation parameters to be aligned with the shapes of objects, ensuring proper decomposition of motions.
Furthermore, our deformation model provides hierarchical structures of primitives for motions, allowing manipulation in a coarse-to-fine manner.

\section{Proposed Method}
\begin{figure*}[t]
    \centering
    \includegraphics[trim={0 20 0 0pt}, width=1\linewidth]{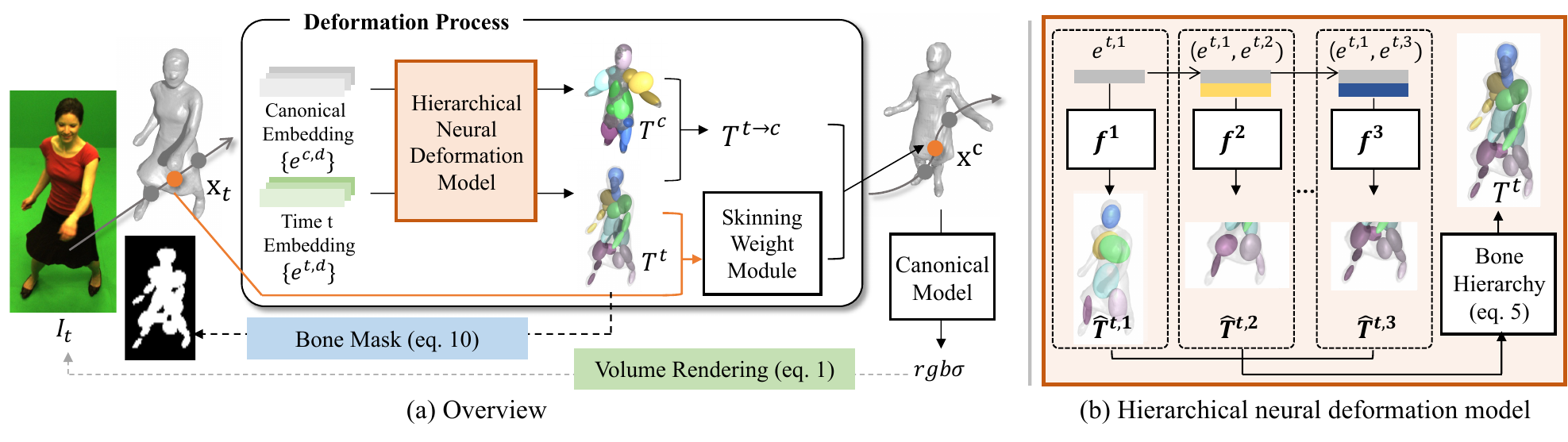}
    \caption{
    (a) The overview of the proposed framework for creating 3D animatble models from videos. 
    Each ray from the image pixel is deformed to the canonical space.
    Rays are deformed in a coarse-to-fine manner, using the hierarchical neural deformation model.
    (b) The process of hierarchical neural deformation model. Coarse motions and fine motions are composited through the bone hierarchy formulation.
    }
   \label{fig:method}
\end{figure*}

Our goal is to construct a framework for creating 3D animatable models of articulated objects from casually captured videos, offering structured bones for easier manipulation.
We first deliver preliminaries \cite{yang2022banmo} (\sref{sec:prelim}), and then introduce our key components, hierarchical deformation model (\sref{sec:hier}), and bone occupancy function (\sref{sec:bone}).
The overall process is outlined in \fref{fig:method} (a).
Our method extends the overall framework of BANMo \cite{yang2022banmo}, with a key difference being our hierarchical deformation model and bone occupancy function.

\subsection{Preliminary}
\label{sec:prelim}
BANMo \cite{yang2022banmo} proposes to reconstruct animatable 3D models from RGB videos through the NeRF \cite{mildenhall2021nerf} framework.
It comprises the time-invariant canonical model and the time-variant deformation model, where the deformation is defined by ellipsoidal bones and the neural skinning weight module.
Given monocular RGB videos, these bones are responsible for deforming rays at each frame to the canonical pose.
Then the canonical model represents the shape and the appearance of the deformed rays in the canonical pose.
All components are jointly optimized together through the differentiable volume rendering.

\textblock{Canonical Model} represents the shape and appearance of an object as NeRFs, $g_c: (\textbf{x}^c, \textbf{d}) \rightarrow (\textbf{c}, \sigma)$, which takes 3D point $\textbf{x}^c=(x,y,z)$ in the canonical space and viewing direction $\textbf{d}=(\phi,\theta)$ as inputs, and produces color $\textbf{c}=(r,g,b)$ and density $\sigma$.
Following VolSDF \cite{yariv2021volsdf}, the SDF value $s$ is produced for mesh extraction, then $s$ is transformed into $\sigma$ as $\sigma=\alpha \big( \frac{1}{2}+\frac{1}{2}sgn(-s) \big(1-exp(-\frac{|-s|}{\beta}) \big) \big)$, where $\alpha$ and $\beta$ are learnable parameters.

\textblock{Volume Rendering.}
To render a frame $I_t$ at time $t$, rays $r^t$ are cast from each pixel using a camera projection matrix.
The $i$-th sampled points $\textbf{x}^t_i$ in $r^t$ are deformed to the canonical space as $\textbf{x}^c_i=T^{t \rightarrow c}\textbf{x}^t_i$.
In the canonical space, $c_i$ and $\sigma_i$ of the deformed points $\textbf{x}^c_i$ are queried from the canonical model.
These values are composited to render the color of $r^t$ through the volume rendering:
\begin{equation}
    \hat{C}(r^t) = \sum_{i=1}^{N} \tau_i (1-exp(-\sigma_i \delta_i))c_i,
\end{equation}
where $\tau = exp(-\sum_{j=1}^{i-1}\sigma_j \delta_j$) is the accumulated transmittance and $\delta_i$ is the distance between adjacent samples.
The overall components are optimized by minimizing the differences of colors between rendered frames and given videos.

\subsection{Hierarchical Neural Deformation Model}
\label{sec:hier}
To represent motions with coarse-to-fine granularity, we introduce a hierarchical neural deformation model, as depicted in \fref{fig:method} (b).
It takes time embedding vectors for each frame as input, and produces neural bone hierarchy for the frame.
Neural bone hierarchy defines bones as Gaussian ellipsoids, with parent bones capturing coarse motions at larger regions and child bones capturing finer motions at more specific parts.

To deform a 3D point $\mathbf{x}^t$ to canonical space, we compute poses of the leaf bones of neural bone hierarchy $\mathcal{P}^t=\{T^t_1, ..., T^t_B\}$ at time $t$, where $T_b^t \in SE(3)$ refers composited rigid transformation parameters through the bone hierarchy formulation for the $b$-th bone.
From those parameters, the mappings between $\mathcal{P}^t$ and the canonical poses $\mathcal{P}^c$ are defined as
\begin{equation}
    T_b^{t \rightarrow c}=T_b^c \cdot (T_b^t)^{-1},\ \ \ \  T_b^{c \rightarrow t}=T_b^t \cdot (T_b^c)^{-1}.
\end{equation}
Subsequently, the skinning weight $w(\textbf{x}^t, \mathcal{P}^t)$ of $\textbf{x}^t$ is computed through the skinning weight module.
We define the backward warping matrix $\mathcal{W}_{\mathbf{x}}^{t \rightarrow c}$ from time $t$ to the canonical space by linear blend skinning (LBS) with $w$ and $T$:
\begin{equation}
    \mathcal{W}^{t \rightarrow c} = \sum_{b=1}^{B} w_b(\mathbf{x}^t, \mathcal{P}^t)\cdot T_{b}^{t \rightarrow c},
\end{equation}
where $w_b$ is the $b$-th dimension of $w$.
With the warping field, $\mathbf{x}^t$ is deformed to the canonical space as $\mathbf{x}^c = \mathcal{W}^{t \rightarrow c} \mathbf{x}^t$.
As the rigid transformation $T$ is invertible, we can compute the forward warping matrix from the canonical space to time $t$:
\begin{equation}
    \mathcal{W}^{c \rightarrow t} = \sum_{b=1}^{B} w_b(\mathbf{x}^c, \mathcal{P}^c)\cdot T_{b}^{c \rightarrow t}.
\label{forward_warping}
\end{equation}

\begin{figure}[t]
    \centering
    \includegraphics[trim={0 20 0 0pt}, width=1\linewidth]{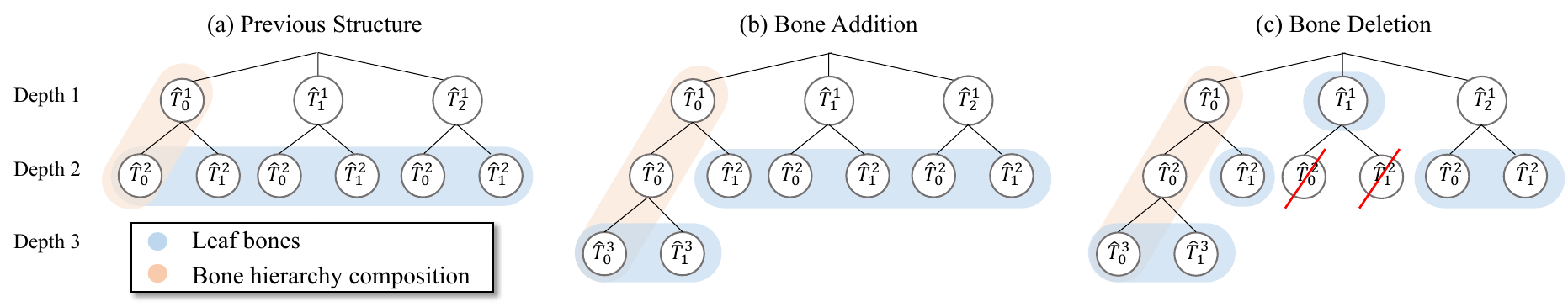}
    \caption{Bone hierarchy diagram. Subordinate bones inherit the motion of all their parent bones (orange line). The leaf bones are used in calculating the skinning weights (blue line). Bones are gradually added during the optimization. After the optimization, users can add or delete the bones in desired regions.}
    \label{fig:tree}
\end{figure}

\textblock{Bone Hierarchy.}
For a structured representation of motions, we organize neural bones in a tree-like structure, where child bones inherit the motions of their parents before making fine-grained movements. 
The diagram of bone hierarchy is depicted in \fref{fig:tree}.
Specifically, for a specific bone at depth $d$, the final transformation $T$ in the world coordinate system is composed by left-multiplying its corresponding parent transformations at previous depths in a recursive way: 
\begin{equation}
    T^{d} = \hat{T}^{1} \hat{T}^{2} \cdots \hat{T}^{d-1} \hat{T}^{d},
\end{equation}
where $\hat{T}^d$ is the local transformation of the bone at depth $d$.
Since the transformations define the center and orient of the bone, this arrangement ensures child bones are defined in the local coordinate system of their parents. 
Starting with a small number of bones at depth 1, as optimization proceeds, each bone is subdivided into child bones of smaller regions with finer-grained motion.

\textblock{Neural Bone Representation.}
We follow a line of previous works \cite{yang2022banmo, yang2021viser, yang2021lasr} and employ 3D Gaussian ellipsoids as the primitives of our bones.
Each bone consists of the rotation $R\in\mathbb{R}^{3\times3}$, the center $\mathbf{t}\in\mathbb{R}^3$ at each time step, and a shared scale vector $\mathbf{s}\in\mathbb{R}^3$ across all time steps.
These are regressed by the MLP $f$ from the embedding vector $e^t$ for each time $t$.
We employ separate MLP $f^d$ for each depth, which takes the embedding of the previous parent bone $e^{t,d}$ and the root embedding $e^{t,1}$ representing global motions.
The local transformation matrix $\hat{T}_i$ of $i$-th bone can be described as
\begin{equation}
    \hat{T}^{t,1}_i, \mathbf{s}^{t,1}_i = f^{1}_i(e^{t,1}), \ \ \ \  \hat{T}^{t,d}_{i}, \mathbf{s}^{t,d}_{i} = f^{d}_{i}([e^{t,1}, e^{t, d-1}]),
\end{equation}
where $f^{d}_{i}$ denotes the $i$-th dimension of the MLP output regressing bones at depth $d$, and $i$ is a local index of the bone within its parent.
The MLP $f^d$ outputs the geometric properties of all child bones at depth $d$.

\textblock{Skinning Weight Module.}
Each point $\mathbf{x}$ is deformed by LBS with the transformation of leaf bones.
The skinning weight of $b$-th leaf bone is defined as
\begin{equation}
    w_b = \frac{exp(-d_M(\mathbf{x}, b) + \Delta w_b)}{\sum_{i=1}^{B} exp( -d_M(\mathbf{x}, b) + \Delta w_b)) },
\end{equation}
\begin{equation}
     d_M(\mathbf{x}, b) =\sqrt{(\mathbf{x}-\mathbf{t}_b)^T R_b^{T}S_bR_b (\mathbf{x}-\mathbf{t}_b)},
\end{equation}
where $d_M(\mathbf{x}, b)$ denotes the mahalanobis distance between $\mathbf{x}$ and $b$-th ellipsoidal bone, and $\Delta w_b$ denotes delta skinning weights computed through MLP, as in \cite{yang2022banmo}.

\textblock{Manipulation.}
With the optimized models, users can manipulate the object into desired poses. To do this, a canonical mesh is extracted by querying the canonical model and applying the marching cube algorithm \cite{loren1987mcube}.
The manipulation of broad movements, which involves the motion of numerous subparts, is achieved by adjusting the parent bones, while finely-tuned motion can be easily achieved by adjusting only the sub-bones.
The canonical mesh is deformed using forward warping in \eref{forward_warping} with the new transformation parameters.

\subsection{Regularizing with Bone Occupancy Function}
\label{sec:bone}
One of the challenges in constructing a bone hierarchy lies in determining the location and the shape of the bones.
Previous work \cite{yang2022banmo} regularizes bone centers using Sinkhorn divergence, yet orients and scales remain under-constrained. 
Consequently, bones are scattered across surfaces and often larger than objects, hindering interpretability and subdivision into finer regions.
To address this challenge, motivated by part-based generative methods \cite{genova2020lif, genova2019sif,paschalidou2021neuralparts}, we propose regularization terms to align the properties of bones (center, orient, scale) with the shape of objects.
The core component of our regularization is the bone occupancy function, which utilizes the mahalanobis distance $d_M(\mathbf{x}, b)$ used in the skinning weight module for identifying the occupancy.

\textblock{Bone occupancy.}
We first model the bone occupancy function $g_b$, which determines the relative position with respect to the surface of bones:
\begin{equation}
    g_b(\mathbf{x}) = d_M(\mathbf{x},b)- \gamma,
\end{equation}
where $\gamma$ is a predefined threshold.
Points inside the bone yield negative values for g(x), while points outside the bone result in positive values.
We further transform $g(x)$ into the density function $\sigma(\frac{-g(x)}{\tau})$, which approximates 1 when $\mathbf{x}$ is inside the bone.
Here, $\sigma$ is a sigmoid function, and $\tau$ is a temperature value determining the sharpness of the boundary.
The bone occupancy function provides ways to relate the locations of the bones with the shapes of the objects. 

\textblock{Bone mask.}
To determine whether a 3D point $\mathbf{x}$ is inside any bones, we define a unified bone occupancy function $G(\mathbf{x})$ by aggregating $g_b(\mathbf{x})$ of all bones:
\begin{equation}
    G(\mathbf{x}) = \min_{b\in{1, \dots B}} g_b(\mathbf{x}).
\end{equation}
With the density obtained from $G(\mathbf{x})$, we construct 2D bone masks $M_{bone}$ by accumulating density values along the ray. We compute the bone mask loss by comparing them with object mask $M_{GT}$ as
\begin{equation}
    \mathcal{L}_{bone} = \sum || M_{bone} - M_{GT} ||^2.
\end{equation}
By regularizing through the bone mask loss, we constrain the location and shape of bones to align with the actual shape of the objects.

\textblock{Overlap \& coverage loss.} We further regularize the properties of bones based on the bone occupancy function. 
We extract surface points $\mathcal{V}$ of the canonical model $g_c(\cdot)$ by applying the marching cube algorithm to the output.
From the points in $\mathcal{V}$, we impose an overlap loss, enforcing that each point is occupied by a maximum of $\lambda$ number of bones:
\begin{equation}
    \mathcal{L}_{overlap} = \frac{1}{|\mathcal{V}|} \sum_{\mathbf{x}\in \mathcal{V}} max\Big(0, \sum_{b=1}^{B} \sigma \big(\frac{-g_b(x)}{\tau} \big) - \lambda \Big).
\end{equation}
In addition, we apply a coverage loss to ensure that each bone occupies a certain portion of the entire region:
\begin{equation}
    \mathcal{L}_{cover} = \sum_{b=1}^{B} \sum_{\mathbf{x} \in \mathcal{N}} \big( max\big( 0, g_b(x) \big) \big),
\end{equation}
where $\mathcal{N}$ denotes the N closest points among $\mathcal{V}$ with respect to mahalanobis distance $d_M(\mathbf{x}, b)$ to the bone.

\subsection{Optimization}\label{sec:optimization}
Our overall system is optimized on given monocular RGB videos, including 2D masks, optical flows, and dense-CSE features extracted from them.
We compute the reconstruction loss term $\mathcal{L}_{recon}$ and cycle loss term $\mathcal{L}_{cycle}$ in BANMo \cite{yang2022banmo}, incorporating additional loss terms related to bones:
\begin{equation}
    \mathcal{L} = \mathcal{L}_{recon} + \mathcal{L}_{cycle} + \mathcal{L}_{bone} + \mathcal{L}_{overlap} + \mathcal{L}_{cover}.
\end{equation}
We refer to Supplement for a more detailed description of $\mathcal{L}_{recon}$ and $\mathcal{L}_{cycle}$.

\textblock{Coarse-to-fine motion optimization.}
To optimize the hierarchical neural deformation system, we propose a coarse-to-fine motion optimization scheme.
We initially optimize depth-1 bones that are responsible for coarse motion with larger region.
During the optimization, we gradually add child bones to the previous bones to progressively capture fine motions.

\textblock{Implementation details.} 
In the optimization process, we start with five initial bones for animals and six for humans. 
After establishing the initial set of bones (parent bones), two additional bones (child bones) are added to each of the existing bones in subsequent stages.
The optimization for each depth involves 20k iterations.
We use two NVIDIA GeForce RTX 3090 GPUs for the optimization, and each stage takes less than 3 hours in our environment.
Please refer Supplement to more implementation details.

\section{Experiment}

\subsection{Experimental Setup}
\textblock{Datasets.}
We evaluate our method on objects with diverse categories, including humans and animals.
AMA haman dataset~\cite{vlasic2008ama} includes multi-view videos capturing actor performances. We use Swing and Samba sequences for our evaluation on humans, and treat them as monocular videos.
We also use Eagle and Cat data from BANMo dataset~\cite{yang2022banmo} for animals.
Eagle contains videos that are rendered with an animated 3D eagle model, while Cat contains casually captured monocular videos.
In the preprocessing phase, we utilize off-the-shelf models, specifically PointRend~\cite{kirillov2020pointrend}, VCN-robust~\cite{yang2019vcn-robust}, and CSE~\cite{neverova2020cse}, to extract object masks, optical flow, and CSE features.
We employ the videos of Swing, Samba, and Eagle for quantitative evaluation by comparing them to the ground-truth 3D mesh.
We provide more descriptions of datasets and results of diverse animal species in Supplement.

\begin{table}[t]
\setlength{\tabcolsep}{5pt}
\centering
\small
\caption{Quantitative results on Eagle and AMA. * indicates methods that utilize predefined skeletons for optimization. (r) indicates reproduced results. }
\resizebox{\columnwidth}{!}{
\begin{tabular}{c|cc|cc|cc|cc|cc|cc}
\toprule
\multirow{2}{*}{Method} & \multicolumn{2}{c|}{ViSER} & \multicolumn{2}{c|}{BANMo} & \multicolumn{2}{c|}{BANMo(r)} & \multicolumn{2}{c|}{CAMM*} & \multicolumn{2}{c|}{RAC*} & \multicolumn{2}{c}{Ours} \\ \cline{2-13} 
    & CD  & F2  & CD  & F2  & CD  & F2  & CD  & F2  & CD  & F2  & CD  & F2   \\ \hline
Eagle & 19.22 & 24.76  & 8.1  & 56.7  & 4.66  & \underline{81.44}  & \textbf{4.50}        & 81.21    & -      & -     & \underline{4.64}  & \textbf{81.59}       \\
Swing & 16.29        & 19.95       & 9.1         & 57.0         & 7.33          & 64.88         & 9.02        & 56.00       & \textbf{6.10}  & \textbf{70.33}       & \underline{7.11}    & \underline{65.88}       \\
Samba   & 23.28        & 22.47       & -           & -            & 7.22          & 64.99         & 7.50        & 62.17       & \underline{6.63}   & \underline{67.71}       & \textbf{6.15}      & \textbf{72.07}      \\ 
\bottomrule
\end{tabular}
}
\label{table:main}
\end{table}
\begin{figure*}[t]
    \centering
    \includegraphics[trim={0 30 0 0pt}, width=1\linewidth]{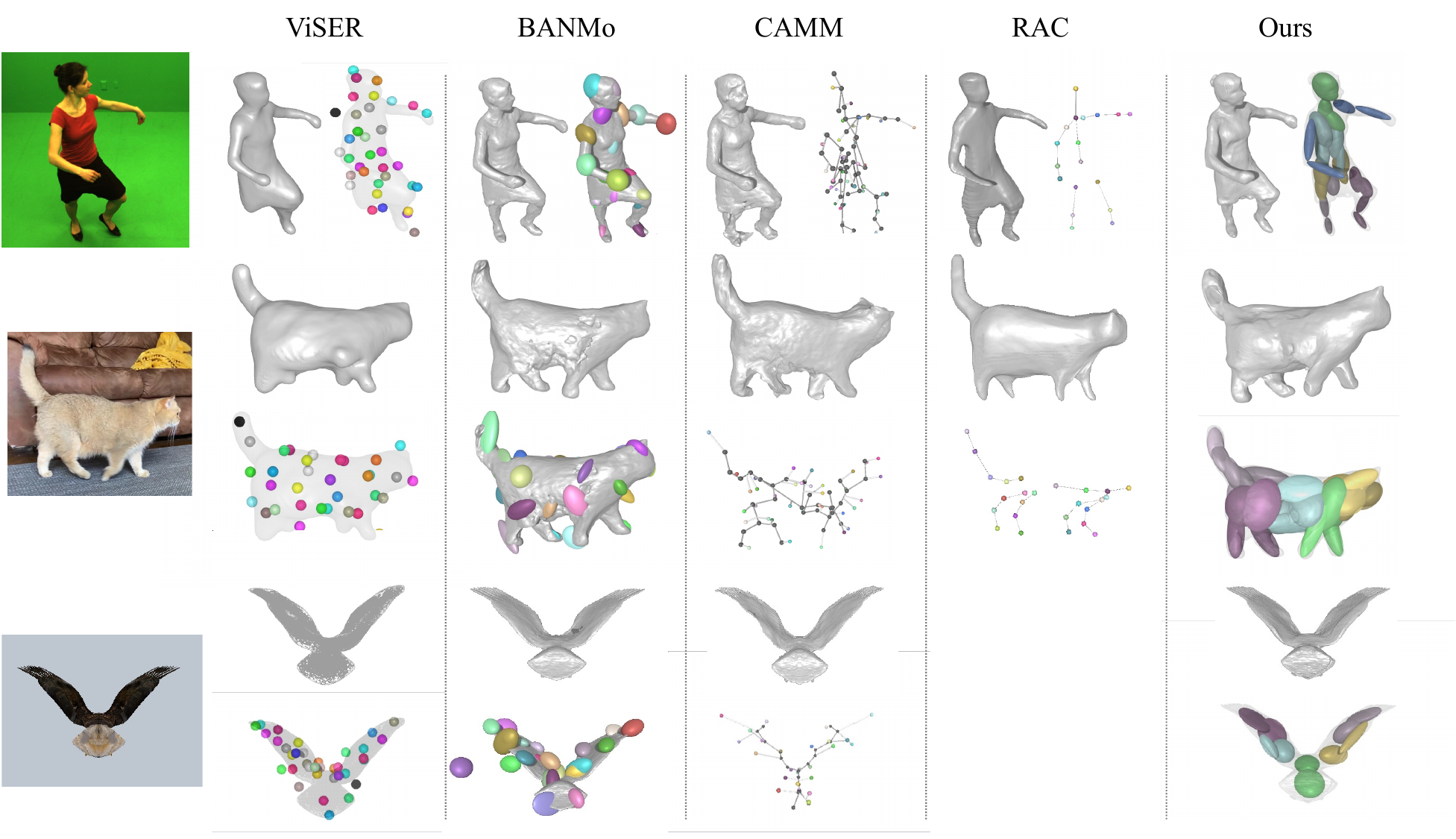}
    \caption{
    Qualitative comparisons with template-free methods (ViSER, BANMo) and skeleton-based methods (CAMM, RAC). 
    The 3D reconstruction results and the corresponding control points are described.
    We omit the eagle result for RAC as they require skeletons for reconstruction, which are not provided.
    }
   \label{fig:results}
\end{figure*}

\textblock{Metrics.} We evaluate the quality of reconstructed 3D objects with the following criteria.
Chamfer Distance (CD) measures the average distance between the ground truth mesh and the estimated surface points. 
We additionally measure F-score at distance thresholds $d=2\%$ (F2) of the longest edge of the axis-aligned object bounding box.
Due to the scale ambiguity, we align the estimated 3D mesh to the ground-truth mesh using Iterative Closest Point before evaluation.

\textblock{Baselines.}
We compare our results with both template-free methods~\cite{yang2021viser, yang2022banmo} and skeleton-based method~\cite{kuai2023camm, yang2023rac}.
\textbf{ViSER}~\cite{yang2021viser} reconstructs 3D articulated objects by learning deformation parameters guided by video-specific surface embeddings.
They utilize 36 ellipsoidal bones for optimization.
\textbf{BANMo}~\cite{yang2022banmo} estimates the pose of the objects using Gaussian ellipsoid bones with canonical NeRF. Total 25 bones are used for all categories of objects.
\textbf{CAMM}~\cite{kuai2023camm} utilizes kinematic chains from RigNet~\cite{RigNet} on top of BANMo to mitigate the challenges associated with manipulating Gaussian bones.
Finally, \textbf{RAC}~\cite{yang2023rac} reconstructs category-level 3D models. 
RAC uses pre-defined skeleton and learns to capture video-specific morphology from videos of diverse instances within the same category. 

\subsection{3D Reconstruction}
\textblock{Quantitative comparison.}
We first quantitatively evaluate the 3D reconstruction results for objects across various categories.
For fair comparisons, we also provide the reproduced results of BANMo as well as the original results reported in their paper.
Due to the absence of the skeleton for eagles, results of RAC on Eagle are omitted.
As shown in \Tref{table:main}, our approach outperforms all template-free methods across all datasets.
We also achieve comparable results with skeleton-based baselines without exploiting predefined structural knowledge.
It is worth noting that our method achieves comparable or better results on Eagle using fewer control points compared to other baselines.
Our method uses only 10 leaf bones for Eagle, whereas other baselines use 25 or more bones to represent deformation.
This demonstrates the efficacy of our structured deformation model in capturing motions with reduced control points, achieving compelling results and potentially improving manipulation interfaces for users.

\textblock{Qualitative comparison.}
\fref{fig:results} describes 3D reconstruction results on Samba, Cat, and Eagle datasets.
Our method accurately reconstructs the 3D models with details.
ViSER shows over-smoothed results with inaccurate poses, which can be attributed to their explicit meshes as shape model and the lack of the ability to aggregate multiple videos.
Methods exploiting NeRF and multiple videos, on the other hand, achieve compelling reconstruction results. 
Methods leveraging predefined skeletons for deformations (RAC and CAMM) generally perform well in capturing poses. 
However, they have difficulty in accurately representing fine details of the motions which are absent in their templates, \eg skirts of the Samba dataset.
We provide more results of such cases in Supplement.

\begin{table}[t]
\setlength{\tabcolsep}{10pt}
\centering
\caption{Quantitative comparison on neural rendering. }
\vspace{-5pt}
\resizebox{0.9\columnwidth}{!}{
\begin{tabular}{c|cc|cc|cc|cc}
\toprule
      & \multicolumn{2}{c|}{Swing} & \multicolumn{2}{c|}{Samba} & \multicolumn{2}{c|}{Eagle} & \multicolumn{2}{c}{Cat}  \\
      & PSNR         & SSIM        & PSNR         & SSIM        & PSNR         & SSIM        & PSNR        & SSIM     \\ \hline
BANMo & 29.53        & 0.921       & 30.72        & 0.916       & 31.05        & 0.900       & 28.01       & 0.850      \\
CAMM & 28.04        & 0.912       & 28.87        & 0.907       & 30.44        & 0.894       & 26.47       & 0.830       \\
RAC &   22.82  &  0.878    &  23.90         &   0.878      & -            & -       &  18.25       &  0.782      \\
Ours  & \textbf{30.43} &  \textbf{0.938} &  \textbf{31.74} &  \textbf{0.942} &  \textbf{32.63} & \textbf{0.924}   & \textbf{28.45}   & \textbf{0.859}      \\ \bottomrule
\end{tabular}
}
\label{table:volume_rendering_main}
\end{table}
\begin{figure*}[t]
    \centering
    \includegraphics[trim={0 35 0 0pt}, width=1\linewidth]{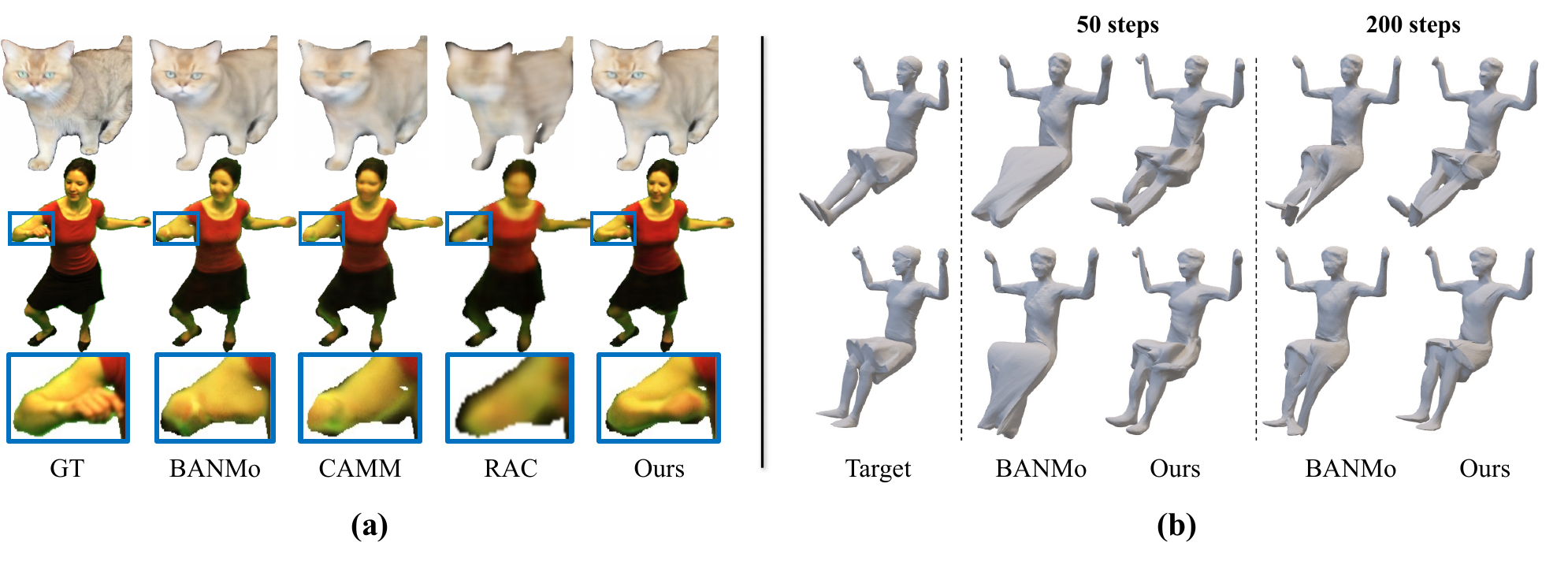}
    \caption{
    (a) Qualitative comparison on neural rendering results. (b) Qualitative comparison of the retargeted objects.
    }
   \label{fig:retargeting}
\end{figure*}

\textblock{Control points comparison.}
To illustrate the interpretability of our framework, we also visualize the control points of various methods in \fref{fig:results}.
For simplicity, our results only visualize leaf bones, with bones sharing parents colored in the same tint.
As can be seen, our bones are aligned within the body, each of which sufficiently covers the parts of the objects.
Our coarse bones capture the parts in a more broader context, \eg the upper body of Samba, the wings of Eagle.
Our child bones in deeper levels sub-divide these coarse parts and represent finer motions at more specific components of the objects.
This can be also clearly seen in \fref{fig:ablation} (a), where bones assigned to the same parent exhibit strong correlations in movements.
In contrast, the resulting control points of BANMo are scattered across the object surfaces without considering the structure and the granularity of motions, resulting in difficulty of understanding and animating the 3D models.
The bone hierarchy of our system provides organized control points for the deformations, enhancing understanding of controls and a more user-friendly manipulation experience.

\subsection{Neural Rendering}
We compare the rendered results with NeRF-based methods.
For quantitative evaluation, we measure the PSNR and SSIM scores between the rendered results and the ground-truth images.
As shown in \Tref{table:volume_rendering_main}, our method outperforms all baselines across diverse categories of objects, demonstrating that our hierarchical modeling of motion enhances rendering quality as well. 
\fref{fig:retargeting} (a) illustrates the rendering results on the Cat and Samba datasets. 
Evident in the detailed motion of the arm (highlighted in the blue box), our method effectively captures intricate movements, resulting in clearer RGB renderings.

\begin{table}[t]
\begin{adjustbox}{valign=t}

\begin{minipage}{0.4\linewidth}
\setlength{\tabcolsep}{6pt}
\centering
\caption{
Quantitative comparison (CD) of the retargeted objects.
}
\label{table:retargeting}
\resizebox{\columnwidth}{!}{
\begin{tabular}{c|cccc}
\toprule
\#steps & 50 & 100 & 150 & 200 \\ \hline
BANMo & 2.75  & 2.03  & 1.90 & 1.86 \\
Ours    & \textbf{2.15} & \textbf{1.93} & \textbf{1.83} &  \textbf{1.75} \\ 
\bottomrule
\end{tabular}
}
\end{minipage}
\end{adjustbox}
\hfill
\begin{adjustbox}{valign=t}
\begin{minipage}{0.59\linewidth}
\setlength{\tabcolsep}{7pt}
\centering
\caption{
Ablation results on the number of videos.
}
\label{table:ablation_frames}
\resizebox{\columnwidth}{!}{
\begin{tabular}{c|cccccc}
\toprule
\multirow{2}{*}{\#videos} & \multicolumn{2}{c}{1 vid} & \multicolumn{2}{c}{4 vids} & \multicolumn{2}{c}{8 vids} \\ \cline{2-7} 
                          & CD        & F2        & CD        & F2        & CD        & F2        \\ \hline
CAMM                      & 17.03     & 38.52     & 10.65     & 48.72     & 7.50      & 62.17     \\
BANMo                     & 10.28     & 47.70     & 11.34     & 45.20     & 7.22      & 64.99     \\
Ours                      & \textbf{9.92}      & \textbf{52.29}     & \textbf{7.05}      &  \textbf{62.34}     & \textbf{6.15}      & \textbf{72.07}     \\ 
\bottomrule
\end{tabular}
}
\end{minipage}
\end{adjustbox}
\end{table}
\begin{table}[t]
\setlength{\tabcolsep}{7pt}
\centering
\small
\caption{Quantitative ablation results on the number of depths and the regularization.}
\resizebox{\columnwidth}{!}{

\begin{tabular}{cc|cc|cc|ccccc}
\toprule
\multicolumn{2}{c|}{Bone reg.}                   & \multicolumn{2}{c|}{No reg.} & \multicolumn{2}{c|}{Sinkhorn} & \multicolumn{5}{c}{Bone occupancy function}                         \\ \hline
\multicolumn{2}{c|}{(\#depths, \#bones)}         & (1, 6)       & (1, 24)       & (1, 6)        & (1, 24)       & (1, 6) & (1, 12) & \multicolumn{1}{c|}{(1, 24)} & (2, 12) & (3, 24) \\ \hline
\multicolumn{1}{c|}{\multirow{2}{*}{Samba}} & CD & 7.66         & 6.84          & 8.56          & 7.17          & 7.65   & 7.21    & \multicolumn{1}{c|}{7.16}    & 6.87    & 6.15    \\
\multicolumn{1}{c|}{}                       & F2 & 61.38        & 67.66         & 57.23         & 65.67         & 61.93  & 63.78   & \multicolumn{1}{c|}{65.41}   & 66.76   & 72.07   \\ \hline
\multicolumn{1}{c|}{\multirow{2}{*}{Swing}} & CD & 8.96         & 8.37          & 9.60          & 8.39          & 9.27   & 9.37    & \multicolumn{1}{c|}{8.83}    & 7.74    & 7.11    \\
\multicolumn{1}{c|}{}                       & F2 & 55.61        & 59.39         & 52.91         & 59.70         & 54.74  & 54.34   & \multicolumn{1}{c|}{58.29}   & 61.64   & 65.88   \\ 
\bottomrule
\end{tabular}

}
\label{table:ablation}
\end{table}

\subsection{Reanimation}
We further compare the reanimation capability and effectiveness of the learned control points against BANMo \cite{yang2022banmo}.
To this end, we conduct optimization-based motion retargeting experiments, following a previous work \cite{wu2022casa}.
Given canonical shapes and corresponding bone parameters, the objective is to retarget the pose of models to a new target pose through bone adjustments.
Specifically, the transform parameters of bones are optimized to minimize CD between predicted and target shapes while preserving fixed canonical shape and skinning weights.
We rig the ground truth mesh of Samba and craft a sequence of 150 frames depicting a novel motion.
We also provide results with various optimization steps (per frame) to illustrate the speed at which we can achieve a target pose.

As shown in \fref{fig:retargeting} (b), we achieve convincing results with fewer optimization steps, thanks to our structured property that moves larger regions with similar motion simultaneously.
As the number of steps increases, the fine details of the poses are further refined.
In contrast, BANMo struggles with handling large motions (e.g., seating, as in the first pose), leading to collapsed body structures.
\Tref{table:retargeting} presents a quantitative comparison of the retargeted objects.
We outperform the baseline at all steps, particularly with a significant margin at a small number of steps, implying better animating capability of our method.

\subsection{Manipulation}
\begin{figure*}[t]
    \centering
    \includegraphics[trim={0 30 0 0pt}, width=1\linewidth]{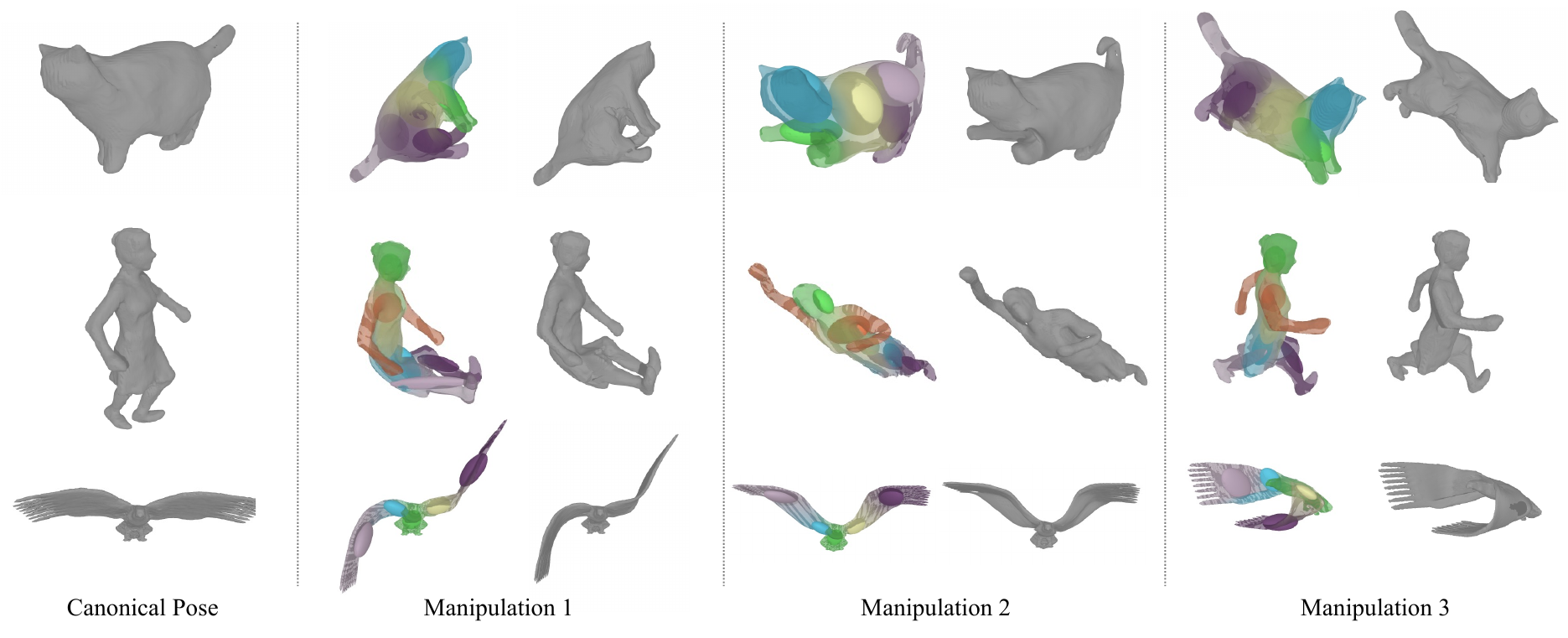}
    \caption{
    Manipulation results on diverse categories of objects. The left side of each column illustrates the depth 1 bones and their corresponding skinning weights, while the right side shows the manipulated results.
    }
   \label{fig:maipulation}
\end{figure*}

We demonstrate the capability of our method in manipulating a diverse set of objects.
The core advantage of our approach is that it provides a coarse-to-fine manipulation, providing easier manipulation for users.
We deliver the example results of the manipulation using our framework in \fref{fig:maipulation}.
Thanks to our tree-structured control points, we can animate various poses with a minimal number of actions.
For instance, we can animate the human and cat to sit using only depth-1 bones (coarsest level), with total 5 movements.
On the other hand, the unstructured bones of BANMo necessitate independent manipulation of the bones to make the same pose, requiring total 25 movements.
In addition, as our deformation model gradually captures the coarse-to-fine structures of the motions, we can flexibly add or delete some of the bones if necessary (\fref{fig:tree}).
If users want to add more control points on the tail of the cat, to better capture detailed motions of it, it can be easily achieved by appending child bones to the corresponding bone.
Note that such dynamic control over the number of bones is not feasible within the framework of BANMo, as its bones lack structure, making it challenging to determine the locations of new bones.
We refer to Supplement for the results of dynamic addition and deletion of the bones.

\begin{figure}[t]
    \centering
    \includegraphics[trim={0 25 0 0pt}, width=1\linewidth]{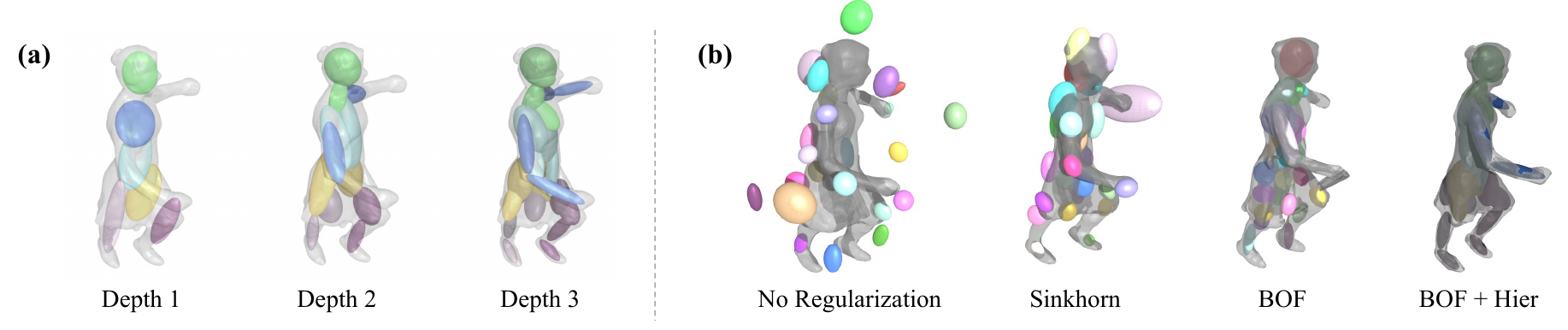}
    \caption{(a) Visualization of the hierarchically structured bones at each depth. (b) Qualitative ablation results on the bone regularization terms.}
    \label{fig:ablation}
\end{figure}

\subsection{Ablation Study}

\textblock{Hierarchical neural deformation model.} 
We ablate our hierarchical neural deformation model by gradually increasing the depths (\#depths = 1, 2, 3).
We compare this to the models without our hierarchy system, which use the same number of bones in one depth (\#bones = 12, 24).
As reported in \Tref{table:ablation}, even when using the same number of bones, the model with our hierarchy system yields much improved quantitative results.
This indicates that capturing coarse motions at the beginning and progressively refining fine-grained movements is more effective in optimizing motions.
Such progressive procedure is more depicted in \fref{fig:ablation} (a).
In the case of Samba, our system first assigns a single bone to the entire leg. As the depth increases, this coarse bone is sub-divided into more specific parts, \eg the calf and the foot, providing correlations between bones with similar motions.

\textblock{Bone Regularization.} 
We then conduct the ablation on the bone regularization terms.
We compare our model with (1) the model without regularization (No regularization) and (2) the model optimized with 
Sinkhorn divergence, as in previous work \cite{yang2022banmo}.
To explore the effects more clearly, we compare these models without our hierarchy system (\#depths = 1).
We deliver the results using 6 and 24 bones, which represent an insufficient and a sufficient number of bones to capture the motions, respectively.
\Tref{table:ablation} and \fref{fig:ablation} (b) present quantitative and qualitative results.
The model regularized with our bone mask loss achieves better results compared to the Sinkhorn divergence loss.
Interestingly, in some cases, the model without regularization delivers the best results.
Despite its quantitative results, as shown in \fref{fig:ablation} (b), bones optimized without any regularization tend to float outside of the body, making it challenging to discern which bone is responsible for a specific part.
The bones regularized with our bone regularization effectively captures motions while being more appropriately placed, achieving improvement when combined with our hierarchy system.

\textblock{Number of input videos.}
Finally, we investigate the performance with a limited number of videos.
We compare the results on Samba by using a single video (1 vid), a half number of videos (4 vids), and all videos (8 vids).
As shown in \Tref{table:ablation_frames}, we outperform baselines in all settings.
BANMo suffers from correctly reconstructing models when using fewer videos, due to the absence of structures in its control points.
On the other hand, our method outperforms BANMo (8 vids) with only using a half number of videos (4 vids), demonstrating the robustness and effectiveness of our structured deformation model.

\section{Discussion and conclusion}
We presented a new framework for creating and animating 3D models, from a set of casually captured videos.
Our hierarchy neural deformation model provides a way to acquire structured bone representations, without exploiting prior structural knowledge, thereby enabling the general applicability of our method.
Combined with the regularization based on the bone occupancy function, our method facilitates easier and interpretable manipulation.
Our approach alleviates the requirements for obtaining animatable models of arbitrary objects, with more comprehensive control points that truly function as ``control points''.

\textblock{Limitation and future works.}
While our structured deformation model provides connections between the bones having similar movements, we expect the motions of the articulated objects can be better captured by the dynamic discovery of joints and conjunction.
Moreover, extending our framework to scenes having multiple objects is a worth exploring subject, which we plan to resolve in our future research.

\section*{Acknowledgements} 
This work was supported by Institute of Information \& communications Technology Planning \& Evaluation (IITP) grant funded by the Korea government(MSIT) (No.2022-0-00124), the National Research Foundation of Korea(NRF) grant funded by the Korea government(MSIT) (NRF- 2022R1A2C2004509), and Institute of Information \& communications
Technology Planning \& Evaluation (IITP) grant funded by the Korea government(MSIT) (Artificial Intelligence Graduate School Program, Yonsei University, under Grant 2020-0-01361).


%
%
\bibliographystyle{splncs04}
\bibliography{main}

\begin{thebibliography}{10}
\providecommand{\url}[1]{\texttt{#1}}
\providecommand{\urlprefix}{URL }
\providecommand{\doi}[1]{https://doi.org/#1}

\bibitem{alldieck2019peopleinclothing}
Alldieck, T., Magnor, M., Bhatnagar, B.L., Theobalt, C., Pons-Moll, G.: Learning to reconstruct people in clothing from a single rgb camera. In: Proceedings of the IEEE/CVF Conference on Computer Vision and Pattern Recognition. pp. 1175--1186 (2019)

\bibitem{alldieck2018humanavatars}
Alldieck, T., Magnor, M., Xu, W., Theobalt, C., Pons-Moll, G.: Detailed human avatars from monocular video. In: 2018 International Conference on 3D Vision (3DV). pp. 98--109. IEEE (2018)

\bibitem{alldieck20183dpeople}
Alldieck, T., Magnor, M., Xu, W., Theobalt, C., Pons-Moll, G.: Video based reconstruction of 3d people models. In: Proceedings of the IEEE Conference on Computer Vision and Pattern Recognition. pp. 8387--8397 (2018)

\bibitem{anguelov2005scape}
Anguelov, D., Srinivasan, P., Koller, D., Thrun, S., Rodgers, J., Davis, J.: Scape: shape completion and animation of people. In: ACM SIGGRAPH 2005 Papers, pp. 408--416 (2005)

\bibitem{blanz2023morphable}
Blanz, V., Vetter, T.: A morphable model for the synthesis of 3d faces. In: Seminal Graphics Papers: Pushing the Boundaries, Volume 2, pp. 157--164 (2023)

\bibitem{bozic2020deepdeform}
Bozic, A., Zollhofer, M., Theobalt, C., Nie{\ss}ner, M.: Deepdeform: Learning non-rigid rgb-d reconstruction with semi-supervised data. In: Proceedings of the IEEE/CVF Conference on Computer Vision and Pattern Recognition. pp. 7002--7012 (2020)

\bibitem{cao2023hexplane}
Cao, A., Johnson, J.: Hexplane: A fast representation for dynamic scenes. In: Proceedings of the IEEE/CVF Conference on Computer Vision and Pattern Recognition. pp. 130--141 (2023)

\bibitem{collet2015high}
Collet, A., Chuang, M., Sweeney, P., Gillett, D., Evseev, D., Calabrese, D., Hoppe, H., Kirk, A., Sullivan, S.: High-quality streamable free-viewpoint video. ACM Transactions on Graphics (ToG)  \textbf{34}(4),  1--13 (2015)

\bibitem{dou2016fusion4d}
Dou, M., Khamis, S., Degtyarev, Y., Davidson, P., Fanello, S.R., Kowdle, A., Escolano, S.O., Rhemann, C., Kim, D., Taylor, J., et~al.: Fusion4d: Real-time performance capture of challenging scenes. ACM Transactions on Graphics (ToG)  \textbf{35}(4),  1--13 (2016)

\bibitem{fridovich2023kplanes}
Fridovich-Keil, S., Meanti, G., Warburg, F.R., Recht, B., Kanazawa, A.: K-planes: Explicit radiance fields in space, time, and appearance. In: Proceedings of the IEEE/CVF Conference on Computer Vision and Pattern Recognition. pp. 12479--12488 (2023)

\bibitem{gao2021dynamic}
Gao, C., Saraf, A., Kopf, J., Huang, J.B.: Dynamic view synthesis from dynamic monocular video. In: Proceedings of the IEEE/CVF International Conference on Computer Vision. pp. 5712--5721 (2021)

\bibitem{genova2020lif}
Genova, K., Cole, F., Sud, A., Sarna, A., Funkhouser, T.: Local deep implicit functions for 3d shape. In: Proceedings of the IEEE/CVF Conference on Computer Vision and Pattern Recognition. pp. 4857--4866 (2020)

\bibitem{genova2019sif}
Genova, K., Cole, F., Vlasic, D., Sarna, A., Freeman, W.T., Funkhouser, T.: Learning shape templates with structured implicit functions. In: Proceedings of the IEEE/CVF International Conference on Computer Vision. pp. 7154--7164 (2019)

\bibitem{hertz2022spaghetti}
Hertz, A., Perel, O., Giryes, R., Sorkine-Hornung, O., Cohen-Or, D.: Spaghetti: Editing implicit shapes through part aware generation. ACM Transactions on Graphics (TOG)  \textbf{41}(4),  1--20 (2022)

\bibitem{innmann2016volumedeform}
Innmann, M., Zollh{\"o}fer, M., Nie{\ss}ner, M., Theobalt, C., Stamminger, M.: Volumedeform: Real-time volumetric non-rigid reconstruction. In: Computer Vision--ECCV 2016: 14th European Conference, Amsterdam, The Netherlands, October 11-14, 2016, Proceedings, Part VIII 14. pp. 362--379. Springer (2016)

\bibitem{kirillov2020pointrend}
Kirillov, A., Wu, Y., He, K., Girshick, R.: Pointrend: Image segmentation as rendering. In: Proceedings of the IEEE/CVF conference on computer vision and pattern recognition. pp. 9799--9808 (2020)

\bibitem{kuai2023camm}
Kuai, T., Karthikeyan, A., Kant, Y., Mirzaei, A., Gilitschenski, I.: Camm: Building category-agnostic and animatable 3d models from monocular videos. In: Proceedings of the IEEE/CVF Conference on Computer Vision and Pattern Recognition. pp. 6586--6596 (2023)

\bibitem{li2024nvfi}
Li, J., Song, Z., Yang, B.: Nvfi: Neural velocity fields for 3d physics learning from dynamic videos. Advances in Neural Information Processing Systems  \textbf{36} (2024)

\bibitem{li2022tava}
Li, R., Tanke, J., Vo, M., Zollh{\"o}fer, M., Gall, J., Kanazawa, A., Lassner, C.: Tava: Template-free animatable volumetric actors. In: European Conference on Computer Vision. pp. 419--436. Springer (2022)

\bibitem{li2022dynerf}
Li, T., Slavcheva, M., Zollhoefer, M., Green, S., Lassner, C., Kim, C., Schmidt, T., Lovegrove, S., Goesele, M., Newcombe, R., et~al.: Neural 3d video synthesis from multi-view video. In: Proceedings of the IEEE/CVF Conference on Computer Vision and Pattern Recognition. pp. 5521--5531 (2022)

\bibitem{li2021nsff}
Li, Z., Niklaus, S., Snavely, N., Wang, O.: Neural scene flow fields for space-time view synthesis of dynamic scenes. In: Proceedings of the IEEE/CVF Conference on Computer Vision and Pattern Recognition. pp. 6498--6508 (2021)

\bibitem{lin2022occlusionfusion}
Lin, W., Zheng, C., Yong, J.H., Xu, F.: Occlusionfusion: Occlusion-aware motion estimation for real-time dynamic 3d reconstruction. In: Proceedings of the IEEE/CVF Conference on Computer Vision and Pattern Recognition. pp. 1736--1745 (2022)

\bibitem{liu2021neuralactor}
Liu, L., Habermann, M., Rudnev, V., Sarkar, K., Gu, J., Theobalt, C.: Neural actor: Neural free-view synthesis of human actors with pose control. ACM transactions on graphics (TOG)  \textbf{40}(6),  1--16 (2021)

\bibitem{loper2023smpl}
Loper, M., Mahmood, N., Romero, J., Pons-Moll, G., Black, M.J.: Smpl: A skinned multi-person linear model. In: Seminal Graphics Papers: Pushing the Boundaries, Volume 2, pp. 851--866 (2023)

\bibitem{loren1987mcube}
Lorensen, W.E., Cline, H.E.: Marching cubes: A high resolution 3d surface construction algorithm. SIGGRAPH Comput. Graph.  (1987)

\bibitem{mildenhall2021nerf}
Mildenhall, B., Srinivasan, P.P., Tancik, M., Barron, J.T., Ramamoorthi, R., Ng, R.: Nerf: Representing scenes as neural radiance fields for view synthesis. Communications of the ACM  \textbf{65}(1),  99--106 (2021)

\bibitem{neverova2020cse}
Neverova, N., Novotny, D., Szafraniec, M., Khalidov, V., Labatut, P., Vedaldi, A.: Continuous surface embeddings. Advances in Neural Information Processing Systems  \textbf{33},  17258--17270 (2020)

\bibitem{noguchi2022watch}
Noguchi, A., Iqbal, U., Tremblay, J., Harada, T., Gallo, O.: Watch it move: Unsupervised discovery of 3d joints for re-posing of articulated objects. In: Proceedings of the IEEE/CVF Conference on Computer Vision and Pattern Recognition. pp. 3677--3687 (2022)

\bibitem{park2021nerfies}
Park, K., Sinha, U., Barron, J.T., Bouaziz, S., Goldman, D.B., Seitz, S.M., Martin-Brualla, R.: Nerfies: Deformable neural radiance fields. In: Proceedings of the IEEE/CVF International Conference on Computer Vision. pp. 5865--5874 (2021)

\bibitem{park2021hypernerf}
Park, K., Sinha, U., Hedman, P., Barron, J.T., Bouaziz, S., Goldman, D.B., Martin-Brualla, R., Seitz, S.M.: Hypernerf: A higher-dimensional representation for topologically varying neural radiance fields. ACM Trans. Graph.  \textbf{40}(6) (dec 2021)

\bibitem{paschalidou2021neuralparts}
Paschalidou, D., Katharopoulos, A., Geiger, A., Fidler, S.: Neural parts: Learning expressive 3d shape abstractions with invertible neural networks. In: Proceedings of the IEEE/CVF Conference on Computer Vision and Pattern Recognition. pp. 3204--3215 (2021)

\bibitem{pavlakos2019expressive}
Pavlakos, G., Choutas, V., Ghorbani, N., Bolkart, T., Osman, A.A., Tzionas, D., Black, M.J.: Expressive body capture: 3d hands, face, and body from a single image. In: Proceedings of the IEEE/CVF conference on computer vision and pattern recognition. pp. 10975--10985 (2019)

\bibitem{peng2021animatable-nerf}
Peng, S., Dong, J., Wang, Q., Zhang, S., Shuai, Q., Zhou, X., Bao, H.: Animatable neural radiance fields for modeling dynamic human bodies. In: Proceedings of the IEEE/CVF International Conference on Computer Vision. pp. 14314--14323 (2021)

\bibitem{peng2021neuralbody}
Peng, S., Zhang, Y., Xu, Y., Wang, Q., Shuai, Q., Bao, H., Zhou, X.: Neural body: Implicit neural representations with structured latent codes for novel view synthesis of dynamic humans. In: Proceedings of the IEEE/CVF Conference on Computer Vision and Pattern Recognition. pp. 9054--9063 (2021)

\bibitem{pumarola2021d-nerf}
Pumarola, A., Corona, E., Pons-Moll, G., Moreno-Noguer, F.: D-nerf: Neural radiance fields for dynamic scenes. In: Proceedings of the IEEE/CVF Conference on Computer Vision and Pattern Recognition. pp. 10318--10327 (2021)

\bibitem{su2021a-nerf}
Su, S.Y., Yu, F., Zollh{\"o}fer, M., Rhodin, H.: A-nerf: Articulated neural radiance fields for learning human shape, appearance, and pose. Advances in Neural Information Processing Systems  \textbf{34},  12278--12291 (2021)

\bibitem{tertikas2023partnerf}
Tertikas, K., Despoina, P., Pan, B., Park, J.J., Uy, M.A., Emiris, I., Avrithis, Y., Guibas, L.: Partnerf: Generating part-aware editable 3d shapes without 3d supervision. arXiv preprint arXiv:2303.09554  (2023)

\bibitem{tretschk2021nr-nerf}
Tretschk, E., Tewari, A., Golyanik, V., Zollh{\"o}fer, M., Lassner, C., Theobalt, C.: Non-rigid neural radiance fields: Reconstruction and novel view synthesis of a dynamic scene from monocular video. In: Proceedings of the IEEE/CVF International Conference on Computer Vision. pp. 12959--12970 (2021)

\bibitem{tu2023dreamo}
Tu, T., Li, M.F., Lin, C.H., Cheng, Y.C., Sun, M., Yang, M.H.: Dreamo: Articulated 3d reconstruction from a single casual video. arXiv preprint arXiv:2312.02617  (2023)

\bibitem{vlasic2008ama}
Vlasic, D., Baran, I., Matusik, W., Popovi{\'c}, J.: Articulated mesh animation from multi-view silhouettes. In: Acm Siggraph 2008 papers, pp.~1--9 (2008)

\bibitem{wang2023animatabledreamer}
Wang, X., Wang, Y., Ye, J., Wang, Z., Sun, F., Liu, P., Wang, L., Sun, K., Wang, X., He, B.: Animatabledreamer: Text-guided non-rigid 3d model generation and reconstruction with canonical score distillation. arXiv preprint arXiv:2312.03795  (2023)

\bibitem{wang2023root}
Wang, Y., Dong, Y., Sun, F., Yang, X.: Root pose decomposition towards generic non-rigid 3d reconstruction with monocular videos. In: Proceedings of the IEEE/CVF International Conference on Computer Vision. pp. 13890--13900 (2023)

\bibitem{weng2022humannerf}
Weng, C.Y., Curless, B., Srinivasan, P.P., Barron, J.T., Kemelmacher-Shlizerman, I.: Humannerf: Free-viewpoint rendering of moving people from monocular video. In: Proceedings of the IEEE/CVF conference on computer vision and pattern Recognition. pp. 16210--16220 (2022)

\bibitem{wu2022casa}
Wu, Y., Chen, Z., Liu, S., Ren, Z., Wang, S.: Casa: Category-agnostic skeletal animal reconstruction. Advances in Neural Information Processing Systems  \textbf{35},  28559--28574 (2022)

\bibitem{xian2021spacetime}
Xian, W., Huang, J.B., Kopf, J., Kim, C.: Space-time neural irradiance fields for free-viewpoint video. In: Proceedings of the IEEE/CVF Conference on Computer Vision and Pattern Recognition. pp. 9421--9431 (2021)

\bibitem{xiang2019monocular}
Xiang, D., Joo, H., Sheikh, Y.: Monocular total capture: Posing face, body, and hands in the wild. In: Proceedings of the IEEE/CVF conference on computer vision and pattern recognition. pp. 10965--10974 (2019)

\bibitem{RigNet}
Xu, Z., Zhou, Y., Kalogerakis, E., Landreth, C., Singh, K.: Rignet: Neural rigging for articulated characters. ACM Trans. on Graphics  \textbf{39} (2020)

\bibitem{yang2019vcn-robust}
Yang, G., Ramanan, D.: Volumetric correspondence networks for optical flow. Advances in neural information processing systems  \textbf{32} (2019)

\bibitem{yang2021lasr}
Yang, G., Sun, D., Jampani, V., Vlasic, D., Cole, F., Chang, H., Ramanan, D., Freeman, W.T., Liu, C.: Lasr: Learning articulated shape reconstruction from a monocular video. In: Proceedings of the IEEE/CVF Conference on Computer Vision and Pattern Recognition. pp. 15980--15989 (2021)

\bibitem{yang2021viser}
Yang, G., Sun, D., Jampani, V., Vlasic, D., Cole, F., Liu, C., Ramanan, D.: Viser: Video-specific surface embeddings for articulated 3d shape reconstruction. Advances in Neural Information Processing Systems  \textbf{34},  19326--19338 (2021)

\bibitem{yang2022banmo}
Yang, G., Vo, M., Neverova, N., Ramanan, D., Vedaldi, A., Joo, H.: Banmo: Building animatable 3d neural models from many casual videos. In: Proceedings of the IEEE/CVF Conference on Computer Vision and Pattern Recognition. pp. 2863--2873 (2022)

\bibitem{yang2023rac}
Yang, G., Wang, C., Reddy, N.D., Ramanan, D.: Reconstructing animatable categories from videos. In: Proceedings of the IEEE/CVF Conference on Computer Vision and Pattern Recognition. pp. 16995--17005 (2023)

\bibitem{yariv2021volsdf}
Yariv, L., Gu, J., Kasten, Y., Lipman, Y.: Volume rendering of neural implicit surfaces. Advances in Neural Information Processing Systems  \textbf{34},  4805--4815 (2021)

\bibitem{zollhofer2014realtime}
Zollh{\"o}fer, M., Nie{\ss}ner, M., Izadi, S., Rehmann, C., Zach, C., Fisher, M., Wu, C., Fitzgibbon, A., Loop, C., Theobalt, C., et~al.: Real-time non-rigid reconstruction using an rgb-d camera. ACM Transactions on Graphics (ToG)  \textbf{33}(4),  1--12 (2014)

\end{thebibliography}

\clearpage
\title{(Supplementary Material)\\Hierarchically Structured Neural Bones\\for Reconstructing Animatable Objects\\from Casual Videos} 
\authorrunning{S. Jeon et al.}
\titlerunning{(Supplementary) Hierarchically Structured Neural Bones}
\author{}
\institute{}
\maketitle
\setcounter{figure}{7}
\setcounter{table}{5}
\setcounter{equation}{14}
\setcounter{page}{19}

In this supplementary material, we provide additional details, comparisons, and results of our method: 
\begin{itemize}[leftmargin=0.7cm]
    \item Manipulation UI and comparison in \Sref{sec:manipulation_comparison}.
    \item Manipulation user study in \Sref{sec:user}
    \item Descriptions of the datasets in \Sref{sec:dataset}.
    \item Dynamic addition and deletion of bones in \Sref{sec:dynamic}
    \item Details of our method in \Sref{sec:method_detail}.
    \item Additional ablation studies in \Sref{sec:additional_ablation}.
    \item Additional   s reconstruction results in \Sref{sec:additional_recon}.
    \item Additional manipulation results in \Sref{sec:additional_manipulation}.
    \item Discussion on the societal impacts of our method in \Sref{sec:societal}.
\end{itemize}

\begin{figure*}[t]
    \centering
    \includegraphics[width=1\linewidth]{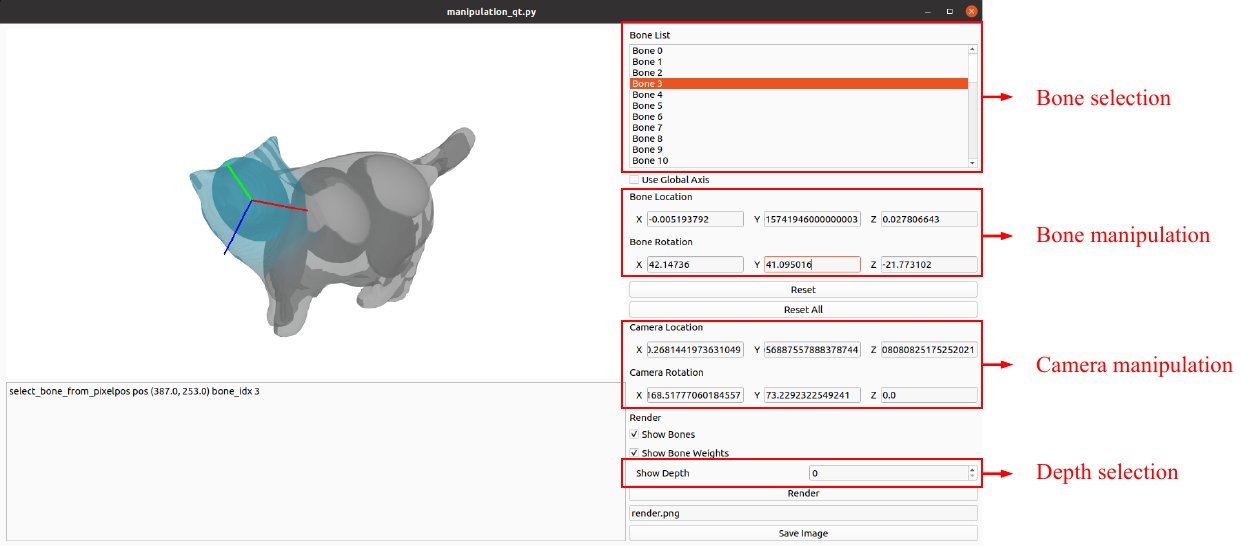}
    \caption{
    Description of our manipulation UI.
    Users can manipulate cameras and bone parameters with mouse actions.
    To select the designated bone, users can see the entire bones or bones at a specific depth, and select the target bone by clicking it in the left side, or choosing it from the bone list on the top-right side.
    The right side shows the manipulation and camera parameters, in which users can directly manipulate these parameters. We refer to the provided supplementary video for more descriptions the actual manipulation process.
    }
   \label{fig:manipulation_ui}
\end{figure*}

\begin{figure}[]
    \centering
    \includegraphics[width=1\linewidth]{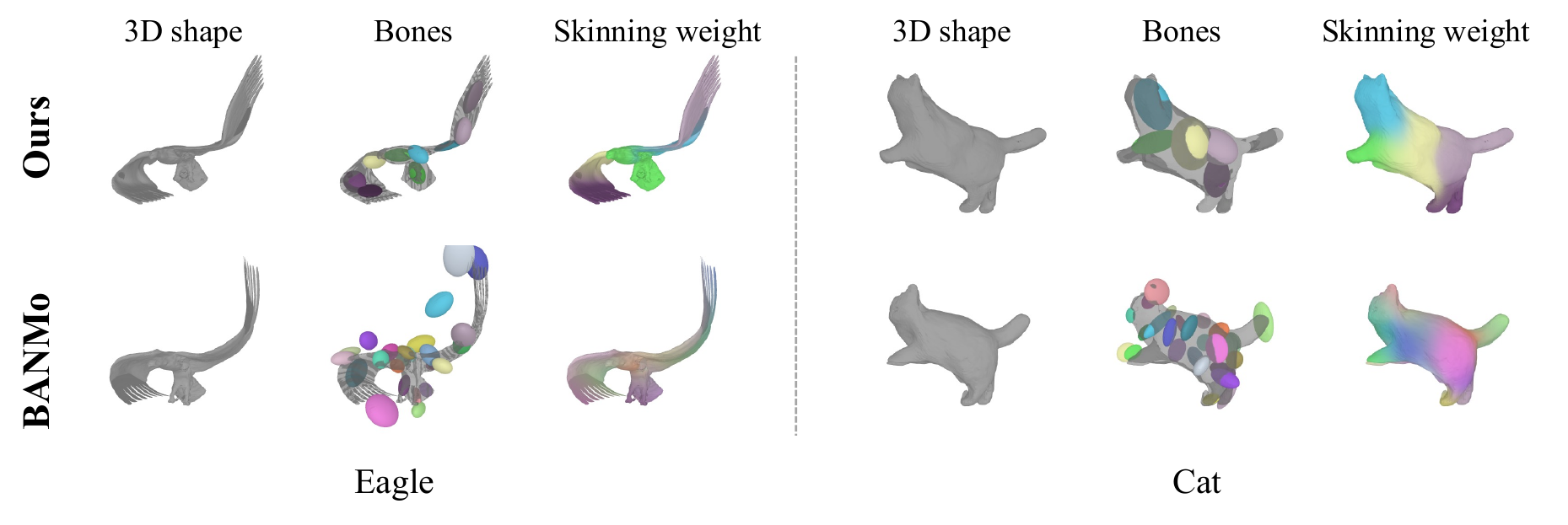}
    \caption{
    Manipulation comparison with BANMo \cite{yang2022banmo}. Users can firstly achieve manipulations of coarse and larger motions using our method, whereas BANMo requires adjustments of almost all bones to make such manipulations. Notably, the manipulation of Eagle is achieved only using 5 bones with our method, while at least 18 bones are adjusted in BANMo.
    }
   \label{fig:manipulation_comparison}
\end{figure}

\section{Manipulation Comparison}
\label{sec:manipulation_comparison}
We showcase the easier and more comprehensible manipulation process achieved by our method through the supplementary videos and manipulated results.
During the manipulation process, the animator utilizes our manipulation UI and manually adjusts the bone parameters to achieve the desired poses of the objects.
We provide a description of our manipulation UI in \fref{fig:manipulation_ui}.
The supplementary video (named \textbf{``manipulation-UI-and-comparison.mp4"}) demonstrates the actual manipulation process of our method and BANMo \cite{yang2022banmo}.
As shown in the video, the manipulation process of our method is much easier and interpretable compared to BANMo, achieving the desired poses in about $4\times$ shorter time.

The manipulated objects are demonstrated in \fref{fig:manipulation_comparison}.
It is worth mentioning that users need to take significantly fewer actions for manipulating our structured deformation model.
For instance, to manipulate Eagle, users can obtain the target pose by manipulating just 5 bones.
In contrast, at least 18 bones are need to be adjusted when manipulating the result of BANMo, as its bones are unstructured, and just scattered throughout the surfaces without considering the basis of motions.
In the manipulation process of Cat, coarse and large motions like standing are achieved by moving coarse-level bones using our method.
On the other hand, the result of BANMo requires adjustments of almost all bones (20 out of 25 bones) to make such manipulations, leading to intricate adjustments and a challenging manipulation process.
Thanks to the hierarchically structured deformation model, the proposed method provides much more intuitive and convenient manipulation process to users.

\begin{figure*}[t]
    \centering
    \includegraphics[width=1\linewidth]{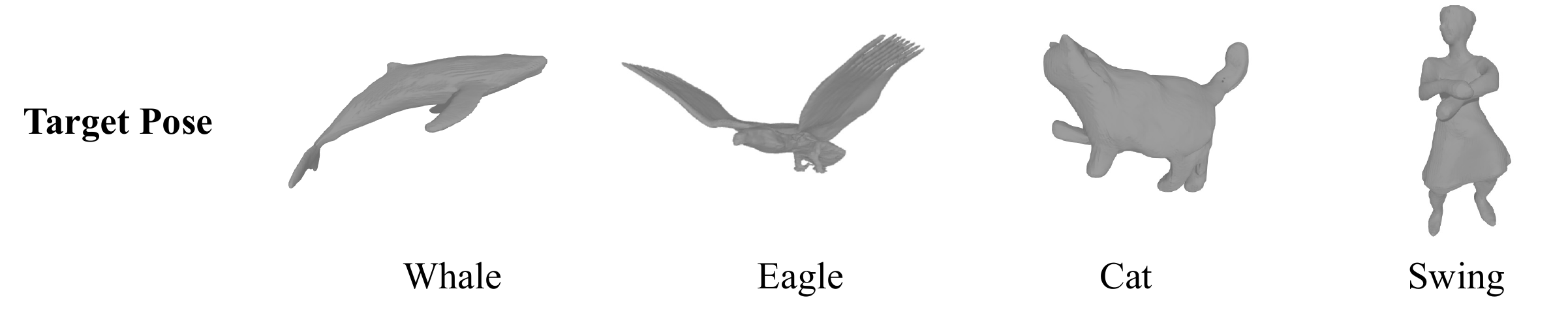}
    \caption{
     The problem presented in the user study. In the user study, we instructed the users to manipulate to achieve the following target pose.
    }
   \label{fig:user_study}
\end{figure*}
\begin{table}[]
\setlength{\tabcolsep}{7pt}
\centering
\caption{User study results on manipulation. }
\resizebox{\columnwidth}{!}{
\begin{tabular}{c|cc|cc|cc|cc|cc}
\toprule
      & \multicolumn{2}{c|}{Whale} & \multicolumn{2}{c|}{Eagle} & \multicolumn{2}{c|}{Cat} & \multicolumn{2}{c|}{Swing} & \multicolumn{2}{c}{Avg.} \\
      & Time         & Pref        & Time         & Pref        & Time        & Pref       & Time         & Pref        & Time        & Pref       \\ \hline
BANMo & 2m 11s         & 3.2           & 3m 48s         & 2.9        & 5m 17s        & 2.6       & 9m 6s        & 1.5        & 5m 5s        & 2.55       \\
Ours  & \textbf{1m 29s}  & \textbf{4.6} &  \textbf{3m 15s} & \textbf{3.7}  & \textbf{3m 31s} &  \textbf{3.9}   &  \textbf{7m 4s}         & \textbf{2.5}  & \textbf{3m 50s}    & \textbf{3.68}       \\ \bottomrule
\end{tabular}}

\label{table:user_study}
\end{table}

\section{Manipulation User Study}
\label{sec:user}
We compare our method with BANMo in terms of manipulation capabilities by conducting a user study. 
For the user study, we recruited 12 participants with no prior experience using 3D tools.
Each participant was instructed to manipulate 3D models to match given target poses.
The test was conducted on four different objects, including Whale, Eagle, Cat, and Swing. 
\fref{fig:user_study} shows the target poses used in the user study.
We measured both the time taken to achieve the desired poses and the preference ratings, rated on a scale from 1 (Difficult) to 5 (Easy).
For each object, we calculated the average of the 10 responses, excluding the shortest and longest times among the 12 responses. 
As shown in \Tref{table:user_study}, our method achieve higher preference ratings and shorter completion times across all objects.
The results demonstrate that our structured bone representation improves manipulation capability in terms of time taken and interpretability of learned control points.

\section{Dataset}
\label{sec:dataset}
We conduct additional experiments on a more diverse range of animals, including a dog, a bat, and a whale:
\begin{itemize}
    \item[-]\textbf{AMA human dataset}~\cite{vlasic2008ama} includes multi-view videos capturing actor performances from 8 synchronized cameras and ground-truth mesh. 
    We select two sets of videos, Swing (1200 frames) and Samba (1400 frames). 
    We omit time synchronization and camera extrinsic parameters during training, treating the videos as monocular.
    \item[-]\textbf{Animated objects dataset}~\cite{yang2022banmo} offers Eagle videos, that are rendered with an animated 3D eagle model and varying camera trajectories.
    Each video comprises 150 frames, and a total of 5 videos are utilized as input.
    \item[-] \textbf{Casual video dataset} \cite{yang2022banmo} includes multiple videos featuring a Cat and a Shiba Inu dog, respectively. 
    These videos are captured casually using monocular cameras, with no control over camera movements. 
    We utilize a total of 11 videos (900 frames) for Cat and 14 videos (1407 frames) for Dog. 
    Specifically, objects exhibit unrestricted movement within individual videos, and the background undergoes changes across the different video sequences.
    \item[-] \textbf{Dynamic object dataset} \cite{li2024nvfi} presents videos of a whale and a bat, which are rendered using animated 3D objects. 
    The animals are depicted from 15 different viewing angles, and for optimization purposes, we utilize videos from 12 of these angles.
    Each video consists of 46 frames, with a total of 552 frames used for both Bat and Whale.
\end{itemize}
\textblock{Dataset license.}
Additionally, we provide the dataset license, the research paper introducing the dataset, and information on whether it includes human subjects in \Tref{table:dataset}.
\begin{table}[t]
\setlength{\tabcolsep}{7pt}
\centering
\small
\caption{Dataset license description. }
\resizebox{\columnwidth}{!}{

\begin{tabular}{llccll}
\toprule
Dataset                 & Instance     & Human & Synthetic & Paper  & License                    \\ \hline
AMA human dataset       & Swing, Samba & \checkmark   &     & \cite{vlasic2008ama} & License not specified      \\
Animated object dataset & Eagle        &       &  \checkmark &  \cite{yang2022banmo} & Turbosquid license         \\
Casual video dataset    & Cat, Dog     &       & &  \cite{yang2022banmo} & CC0                        \\
Dynamic Object dataset  & Bat, Whale   &       & \checkmark &  \cite{li2024nvfi} & SketchFab Standard License \\ \bottomrule
\end{tabular}
}
\label{table:dataset}
\end{table}

\textblock{Human subject.}
We adhere to ethical principles outlined in ECCV ethics guidelines. 
When utilizing human-derived data, particularly in the case of the AMA human dataset, we exercise careful consideration.
The dataset is collected with consent and is made publicly available. 
We utilize the data with proper citation to acknowledge its source.
The dataset is intended for editing purposes, and we ensure its usage aligns with our purpose.
If concerns arise regarding the potential presence of personally identifiable information in facial regions, we pledge to blur or mask the facial area.

\section{Dynamic Addition and Deletion of Bones}
\label{sec:dynamic}
\begin{figure}[t]
    \centering
    \includegraphics[width=1\linewidth]{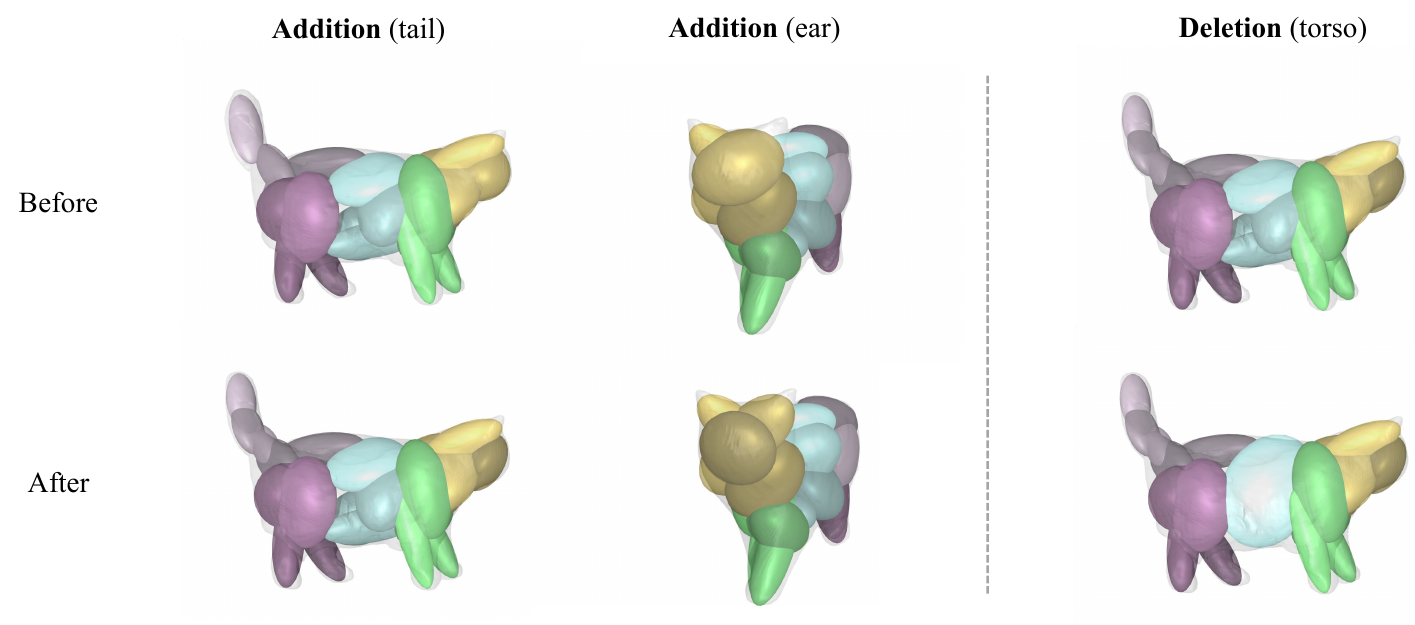}
    \caption{Examples of dynamic addition and deletion of neural bones on the 3D model of Cat. We add extra bones to the tail and head, allowing for manipulation of finer regions. Conversely, the torso, which requires fewer bones, can be merged. }
    \label{fig:cat_dynamic_tree}
\end{figure}

Thanks to the flexible structure of hierarchically structured bones, users have the capability to add additional control points where needed or remove unnecessary ones.
Specifically, users select the designated parent bones to add more bones, and then the child bones are appended to the selected segments accordingly.
With further optimization of the appended bones, users finally obtain the 3D models with more control points for finer manipulation.
For the removal of redundant bones, users select the target bones, and the corresponding child bones can be eliminated by removing them from our tree structures.
This process can be easily implemented by modifying the leaf bones.
We would like to note that prior template-free methods \cite{yang2022banmo, yang2021viser} lack the capability of dynamically adding or removing control points in designated areas, as their Gaussian ellipsoids are unstructured.
Skeleton-based approaches \cite{kuai2023camm, yang2023rac} have insufficient capability of modifying predefined templates, and they offer limited transformations that are restricted to a given skeleton.
\fref{fig:cat_dynamic_tree} illustrates the examples of the dynamic addition and deletion of the bones on Cat.

\section{Method Detail}
\label{sec:method_detail}
\subsection{Losses}
Our method follows reconstruction losses $L_{recon}$ and cycle loss $L_{cycle}$ that are proposed in BANMo~\cite{yang2022banmo}, as follows:
\begin{equation}
    \mathcal{L}_{recon} = \mathcal{L}_{rgb} + \mathcal{L}_{sil} + \mathcal{L}_{OF} + \mathcal{L}_{feat},
\end{equation}
\begin{equation}
    \mathcal{L}_{cycle} = \mathcal{L}_{2D-cyc} + \mathcal{L}_{3D-cyc}.
\end{equation}
\begin{itemize}
    \item \textbf{RGB reconstruction loss $L_{rgb}$} compares rgb values $C_{GT}$ of given frames to the composited values $\hat{C}(r)$, as
    \begin{equation}
        L_{rgb} = \sum_{r} || \hat{C}(r) - C_{GT} ||^2.
    \end{equation}
    
    \item \textbf{Silhouette reconstruction loss $L_{sil}$} compares mask values $M_{GT}$ extracted from given frames and the composited density values $\hat{M}(r)$ through differentiable volume rendering:
    \begin{equation}
        L_{sil} = \sum_{r} || \hat{M}(r) - M_{GT}||^2.
    \end{equation}
    
    \item \textbf{Flow reconstruction loss $L_{OF}$} compares 2D optical flow values $F_{GT}$ extracted from the off-the-shelf flow network and the predicted flow values. In detail, given two frames of time $t$ and $t'$, we compute flows by firstly backward warping rays $r^t$ to the rays in the canonical space $r^c$, then forward warping the rays $r^c$ to the $r^{t'}$ in the $t'$ frame.
    The predicted pixel locations at time $t'$ are compared to the pixel location at time $t$ to compute 2D optical flows $\hat{F}$.
    The flow reconstruction loss is computed as
    \begin{equation}
        L_{OF} = \sum_{r, (t, t')} || \hat{F}\big(r, (t, t')\big) - F_{GT} ||^2.
    \end{equation}

    \item \textbf{Feature rendering loss $L_{feat}$} compares 2D Dense-CSE feature $D_{GT}$ from Dense-CSE~\cite{neverova2020cse} to the composited predicted Dense-CSE feature values $\hat{D}$.
    For each 3D point sampled from rays $r$, the 3D Dense-CSE feature is queried from the feature MLP, and composited to the 2D rendered value.
    \begin{equation}
        L_{feat} = \sum_{r} || \hat{D}(r) - D_{GT} ||^2.
    \end{equation}

    \item \textbf{2D cycle loss $L_{2D-cyc}$} computes cycle consistency between original pixel locations $r$ and the re-projected pixel locations $\hat{r}_{reproj}$. 
    Per each pixel, a 3D point is predicted via canonical embedding in the canonical space.
    The point is warped to time t space (forward warping), and then projected to image space using a predicted camera projection matrix.
    \begin{equation}
        L_{2D-cyc} = \sum_{r} || \hat{r}_{reproj} - r ||^2.
    \end{equation}

    \item \textbf{3D cycle loss $L_{3D-cyc}$} computes cycle consistency of 3D points $\mathbf{x}^t$ by forward warping the canonical points in the canonical space, which was given by backward warping in the time $t$ space as
    \begin{equation}
        L_{3D-cyc} = \sum_{i} \tau || \mathcal{W}^{c \rightarrow t} \cdot \mathcal{W}^{t \rightarrow c} \mathbf{x^t} - \mathbf{x^t}||^2,
    \end{equation}
    where $\tau$ is the opacity of the point $\mathbf{x^t}$.

\end{itemize}

\subsection{Child Bone Initialization}
When increasing the depths of our bone hierarchy, child bones are initialized using properties inherited from their parent bone.
Specifically, a canonical mesh is extracted from the canonical model. 
Skinning weights of previous depths are computed based on the vertices of the canonical mesh. 
We identify vertices with the highest skinning weights on the parent bone and cluster them into groups corresponding to the number of child bones based on euclidean distance. 
The centers of these clusters serve as the initial center positions for the child bones. 
As for the orients of the child bones, we set them to the identity rotation matrix.
For scales, we initialize them with constant values for all bones, regardless of depth.
Using these initial values, the deformation MLP $f^{d}$ for the new depth $d$ is optimized with a small number of iterations.
Since this procedure relies solely on the canonical poses of bones, we discovered that a large number of iterations can lead $f^{d}$ to overfit to these poses. 
Therefore, additional optimization of $f^{d}$ using video data containing various poses is necessary.

\subsection{Additional Optimization Detail}
We optimize our overall system jointly, including the canonical model $g_c$ and the hierarchical deformation model $f$, through the previously mentioned losses.
Specifically, we sample 6 pixels for each image and 128 points are sampled for each ray.
All frames are cropped around the object and resized to the size of $512 \times 512$, and we use 512 images for one iteration.
We use loss weight 1 for $L_{OF}$, $L_{match}$, weight 0.1 for $L_{rgb}, L_{sil}, L_{bone}$, and weight 0.001 for $L_{overlap}$, $L_{cover}$.
As described in the manuscript, we optimize overall system in a coarse-to-fine manner according to the depth of hierarchical neural deformation model.
After parent bones are sufficiently optimized and child bones are appended, we freeze the parent bones and concentrate on optimizing the newly added child bones.
We use two NVIDIA GeForce RTX 3090 GPUs for the optimization, and each stage takes less than 3 hours in our environment.

\begin{figure*}[t]
    \centering
    \includegraphics[width=1\linewidth]{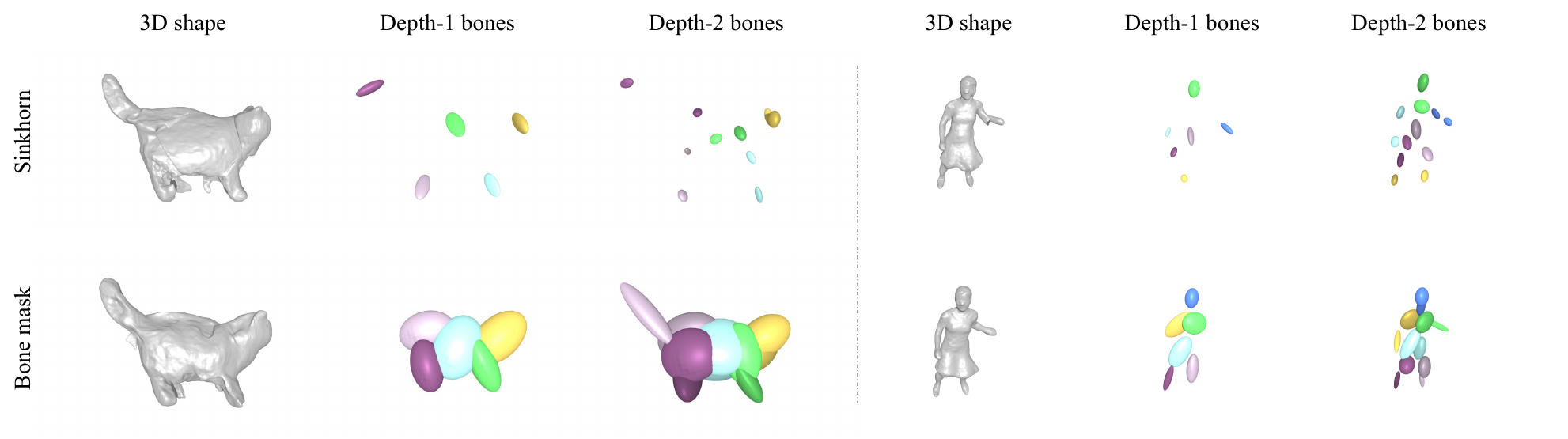}
    \caption{
    Ablation results on the bone regularization terms within the framework of bone hierarchy. When bone mask regularization is utilized, it ensures that the scales of bones correspond to the actual scale of the shape, thereby enabling the subdivision of depth-1 bones into depth-2 bones.
    }
   \label{fig:ablation_sinkhorn}
\end{figure*}

\begin{table}[t]
\setlength{\tabcolsep}{10pt}
\centering
\small
\caption{Ablation on the bone regularization with bone hierarchy. The combination of  bone mask regularization with our hierarchical deformation model achieves the best scores.}
\resizebox{0.9\columnwidth}{!}{
\begin{tabular}{c|cc|cc|cc}
\toprule
\multirow{2}{*}{Bone Reg}  & \multirow{2}{*}{\#depths} & \multirow{2}{*}{\#bones} & \multicolumn{2}{c|}{Samba} & \multicolumn{2}{c}{Swing} \\ \cline{4-7} 
                           &                           &                          & CD        & F2          & CD        & F2        \\ \hline
\multirow{2}{*}{Sinkhorn}  & 1                         & 6                        & 8.56        & 57.23        & 9.60        & 52.91       \\
                           & 2                         & 12                       & 7.84        & 60.79        & 8.88        & 56.22       \\ \hline
\multirow{2}{*}{Bone mask} & 1                         & 6                        & 7.65        & 61.93        & 9.27        & 54.74       \\
                           & 2                         & 12                       & 6.87        & 66.76        & 7.74        & 61.64       \\ 
\bottomrule
\end{tabular}
}
\label{table:supple}
\end{table}

\begin{table}[t]

\setlength{\tabcolsep}{7pt}
\centering
\small
\caption{Progressive optimization ablation on Eagle, Samba, and Swing. BANMo+ extends BANMo by gradually increasing the number of bones during optimization, while BANMo maintains a constant number of bones throughout. Our method also gradually increases the number of bones but utilizes bone hierarchy when adding bones.}
\resizebox{0.7\columnwidth}{!}{
\begin{tabular}{c|cc|cc|cc}
\toprule
\multirow{2}{*}{data} & \multicolumn{2}{c|}{Eagle} & \multicolumn{2}{c|}{Samba} & \multicolumn{2}{c}{Swing} \\ \cline{2-7} 
                      & CD          & F2           & CD          & F2           & CD          & F2          \\ \hline
BANMo                 & 4.66        & 81.44        & 7.22        & 64.99        & 7.33        & 64.88       \\
BANMo+                & 5.52        & 71.61        & 6.82        & 67.17        & 7.13        & 64.56       \\
Ours                  & \textbf{4.64}  & \textbf{81.59}        & \textbf{6.15}        & \textbf{72.07}    & \textbf{7.11}      & \textbf{65.88}       \\ 
\bottomrule
\end{tabular}
}
\label{table:optimization}

\end{table}
\begin{figure*}[t]
    \centering
    \includegraphics[width=1\linewidth]{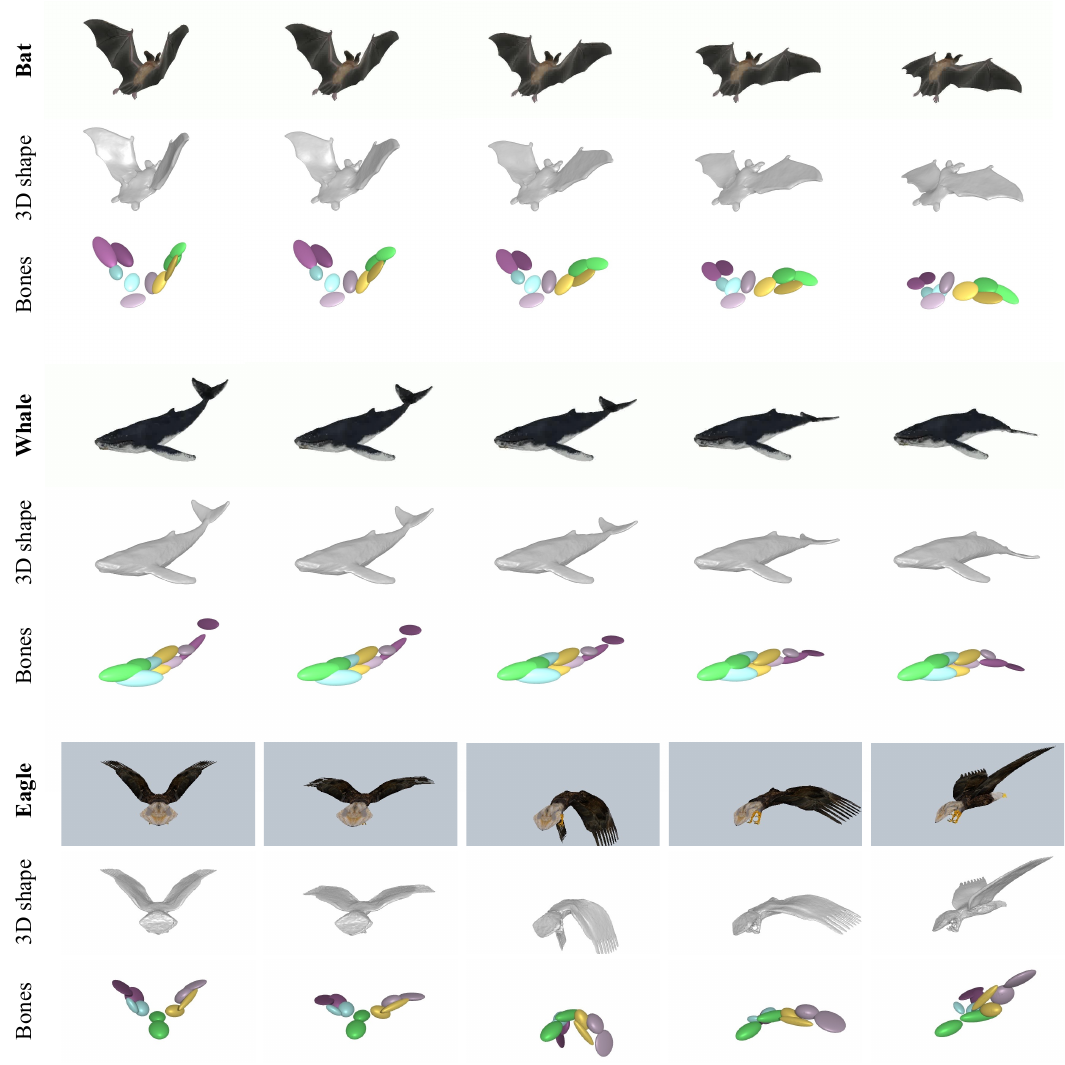}
    \caption{
    Reconstruction results on synthetic animals (Bat, Whale, and Eagle). Reconstructed 3D shapes and their corresponding leaf bones are described.
    }
   \label{fig:supple_animals_syn}
\end{figure*}

\begin{figure*}[t]
    \centering
    \includegraphics[width=1\linewidth]{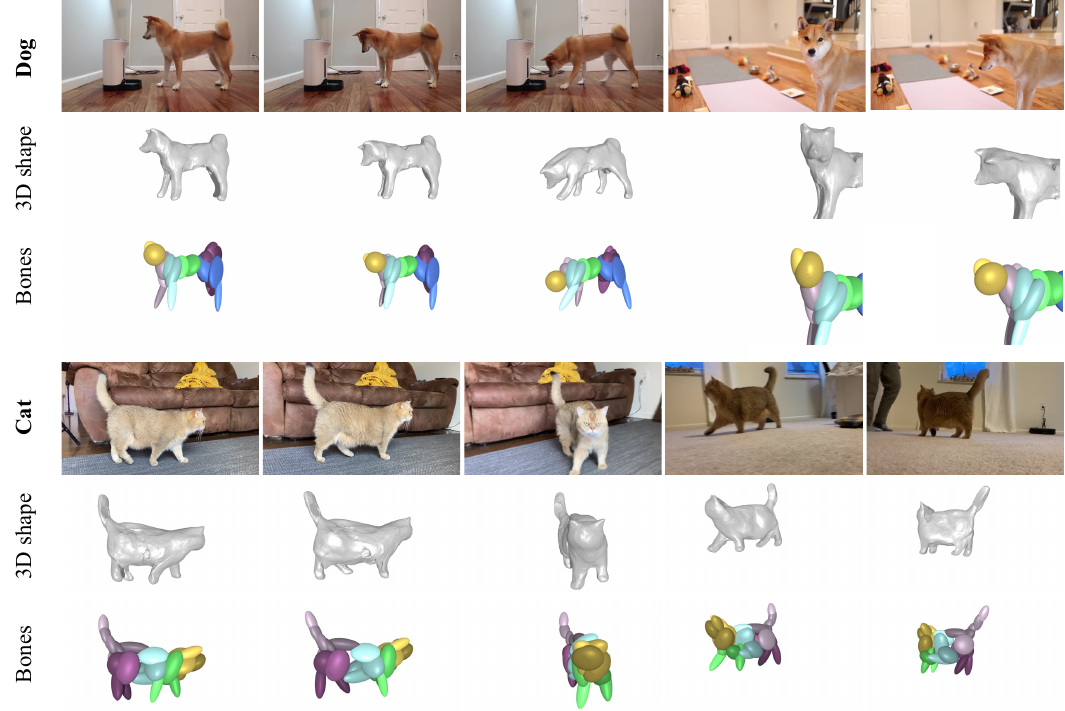}
    \caption{
    Reconstruction results of animals from casually captured videos (Dog, Cat). Reconstructed 3D shapes and their corresponding leaf bones are described.
    }
   \label{fig:supple_animals_real}
\end{figure*}

\begin{figure*}[t]
    \centering
    \includegraphics[width=1\linewidth]{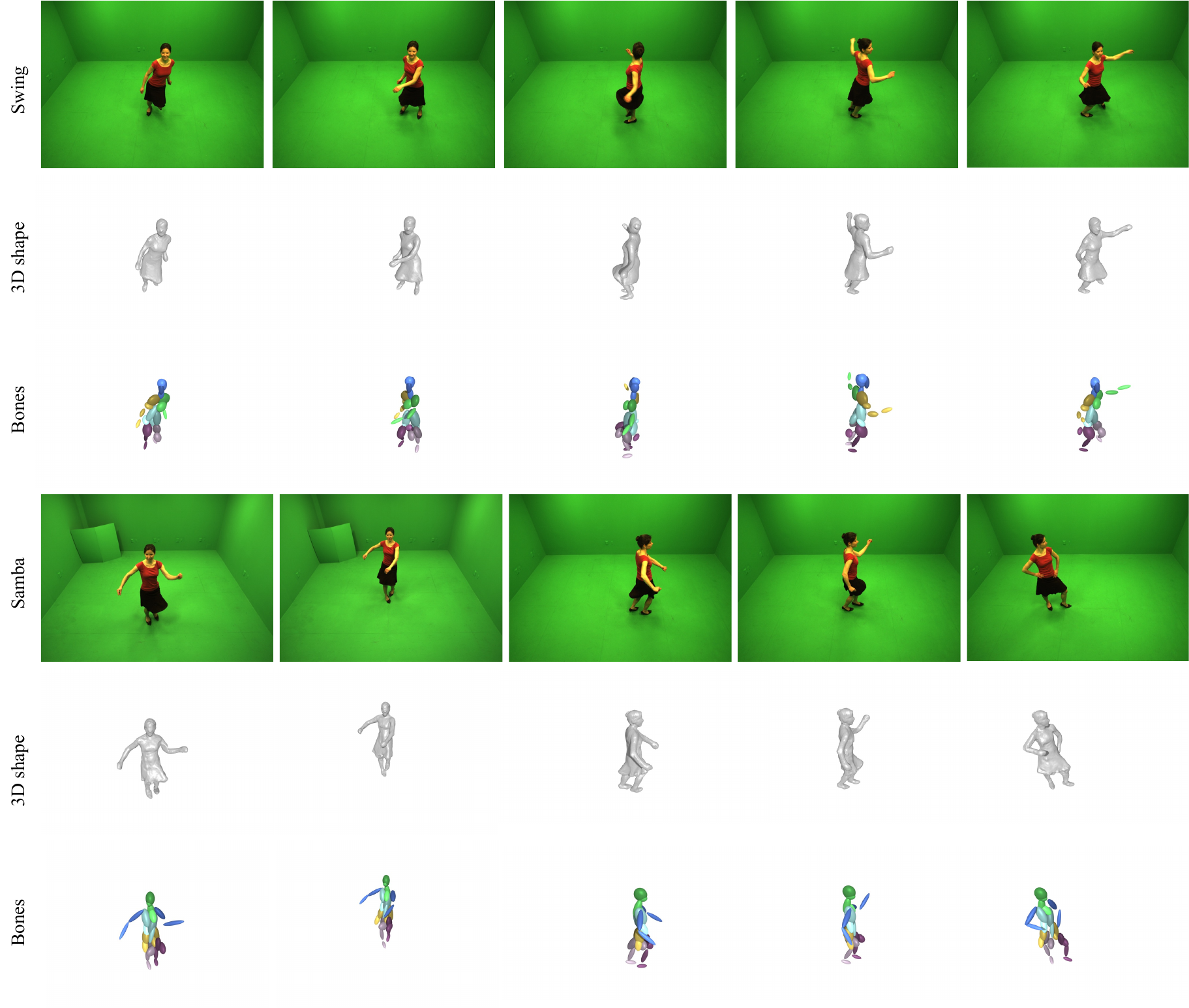}
    \caption{
    Reconstruction results on AMA human datasets (Swing, Samba). Reconstructed 3D shapes and their corresponding leaf bones are described.
    }
   \label{fig:supple_ama}
\end{figure*}

\begin{figure*}[t]
    \centering
    \includegraphics[width=1\linewidth]{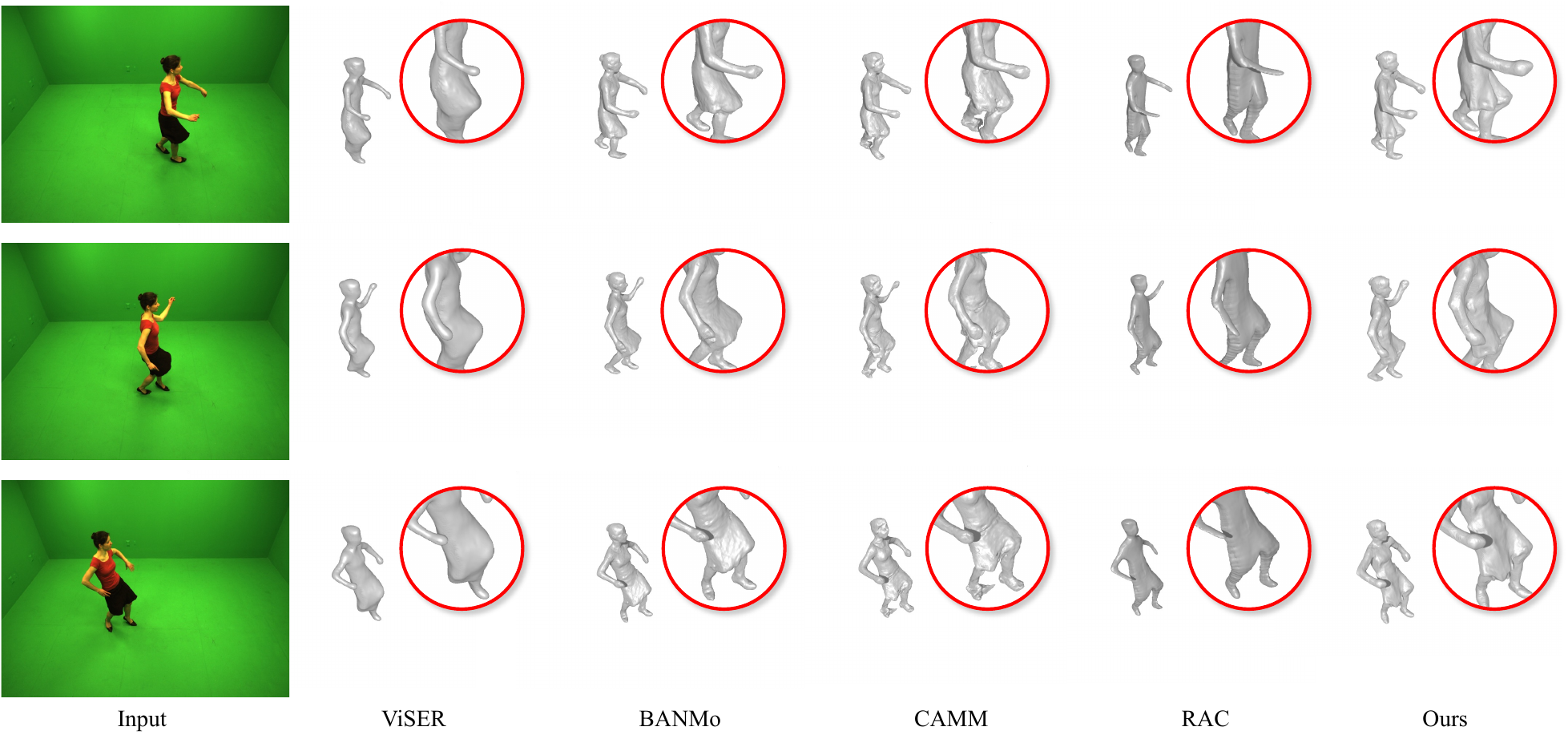}
    \caption{
    Skirt reconstruction of the Samba dataset. Template-free methods excel in reconstructing regions where templates are not provided.
    }
   \label{fig:supple_skirt}
\end{figure*}

\begin{figure*}[t]
    \centering
    \includegraphics[width=1\linewidth]{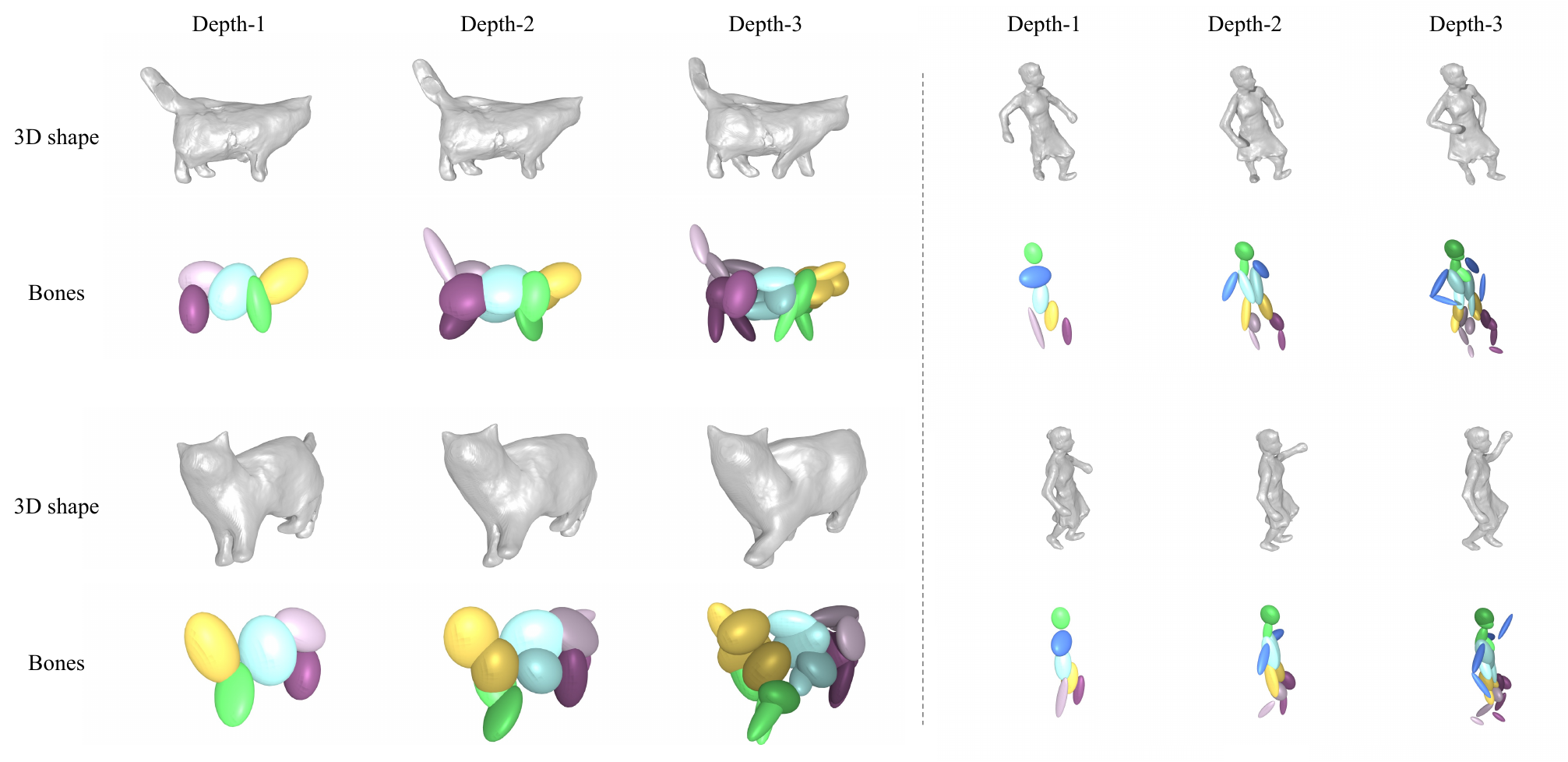}
    \caption{
    Motion specification along depths.
    }
   \label{fig:supple_depth}
\end{figure*}

\section{Additional Ablation Study}
\label{sec:additional_ablation}
\subsection{Bone Regularization}
We further present the ablation results on the effects of combining the bone mask loss with our hierarchical deformation model.
We compare the results at depth-1 and depth-2, with our framework using Sinkhorn divergence regularization as in the prior work \cite{yang2022banmo}.
The reconstructed 3D shapes and their corresponding bones at each depth are reported in \fref{fig:ablation_sinkhorn}.
The most notable difference is that the scale of neural bones align with the scale of the shapes when using bone mask loss.
This effect arises from the fact that Sinkhorn regularization only encourages the center of the bones to be placed near the surfaces, while bone mask loss regularizes all properties of the bones, scales, orients, and centers, by encouraging the bones to fit the foreground masks of the objects.
Combining the bone mask loss with our hierarchical deformation model results in improved interpretability.
Users can better understand the corresponding parts assigned to each bone, while semantic correlations between the bones emerge through the tree structures.
The combination of bone mask loss with our hierarchical deformation model also leads to more notable improvement in reconstruction quality, as can be observed in \Tref{table:supple}.

\subsection{Progressive Optimization}
In our framework, the number of bones increases gradually as depth grows and is further optimized.
To analyze whether the improvement arises from hierarchical modeling or the gradual increase in the number of elements, we conduct additional ablation studies on the optimization process.
For the analysis, we introduce BANMo$+$, in which a small number of bones are initialized and optimized in the initial stage. 
Subsequently, additional bones are progressively added and then re-optimized. 
We begin with 6 bones in the first stage, doubling their quantity over 3 stages, resulting in a total of 24 bones.
We employ identical settings for progressive optimization as in our hierarchical bones.
As shown in \Tref{table:optimization}, BANMo+ does not bring meaningful improvement, sometimes showing degraded results compared to BANMo.
The results suggest that the advancement of our framework is primarily due to the structured modeling of foundational elements, which facilitates the disentanglement of coarse and fine motions.

\section{Additional Reconstruction Result}
\label{sec:additional_recon}
Reconstruction results for a wider range of object categories are illustrated in \fref{fig:supple_animals_syn} and \fref{fig:supple_animals_real}. 
We also present the learned bones, where the bones with the same color indicate the bones assigned to the same parent.
Our method demonstrates generalizability across diverse types of animals with distinct motion properties.
More results of Samba and Swing are depicted in \fref{fig:supple_ama}. 
Template-free methods excel in reconstructing regions where templates are not provided, such as the skirts of humans.
We emphasize and showcase such cases in \fref{fig:supple_skirt}.
Additionally, we present reconstruction results along depths in \fref{fig:supple_depth}.
As the depth increases, the detailed motion \eg legs of the cat, and arms of human, is captured.
For more results and comparisons, please refer to our supplementary video.

\section{Additional Manipulation Result}
\label{sec:additional_manipulation}
\begin{figure*}[t]
    \centering
    \includegraphics[width=1\linewidth]{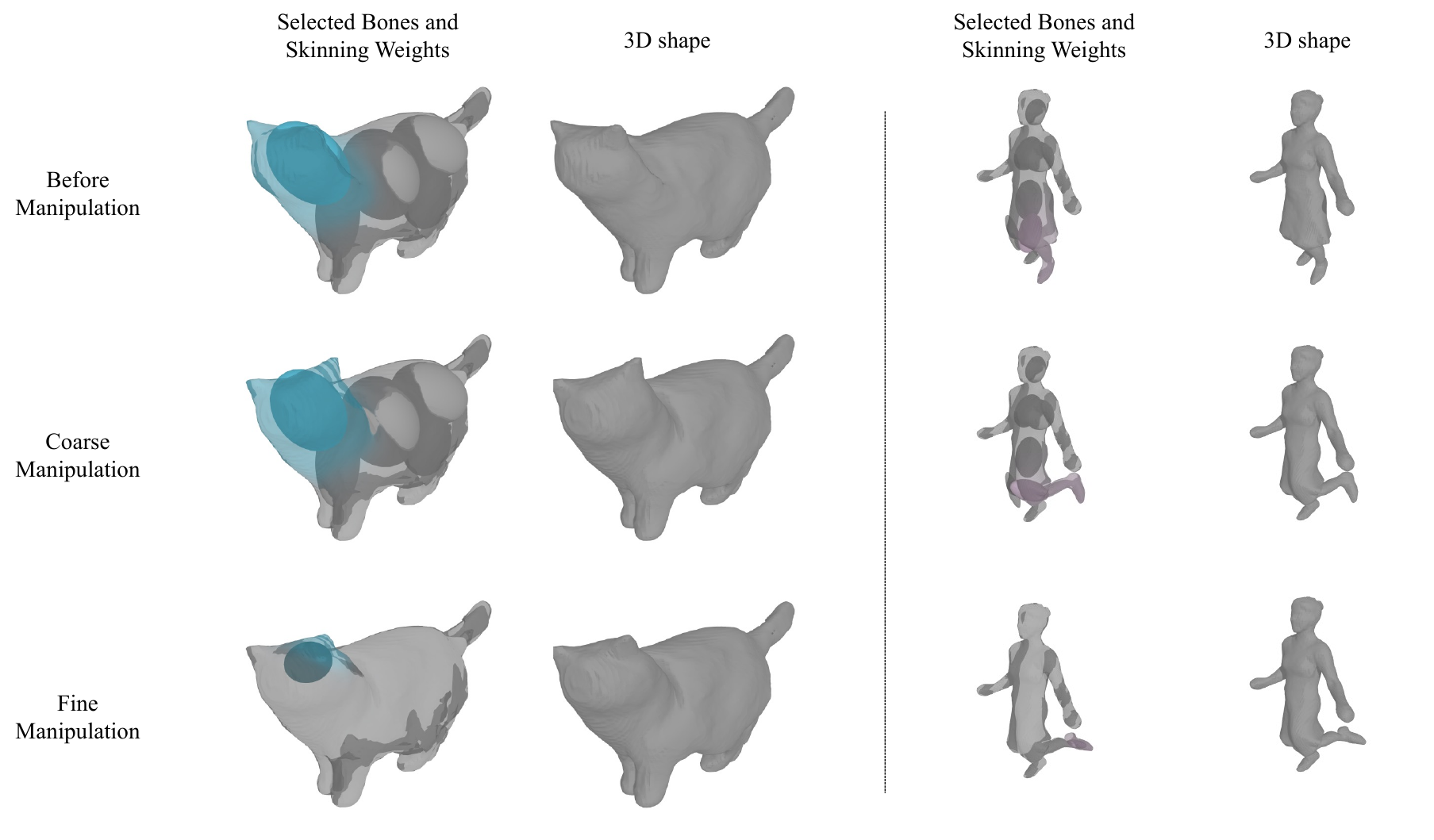}
    \caption{
    Results of coarse-to-fine manipulation.
    }
   \label{fig:coarse_to_fine}
\end{figure*}

\textblock{Coarse-to-fine manipulation. }\fref{fig:coarse_to_fine} outlines the process of coarse-to-fine manipulation employing our hierarchical deformation models. 
In the coarse manipulation, all child bones are adjusted simultaneously, \eg the head of cat and the left leg of the human.
In the fine manipulation, child bones are manipulated in the local coordinate of their parents, enabling the fine adjustments of the motions, as shown in the left ear of the cat, and the foot of the human.

\begin{figure*}[t]
    \centering
    \includegraphics[width=1\linewidth]{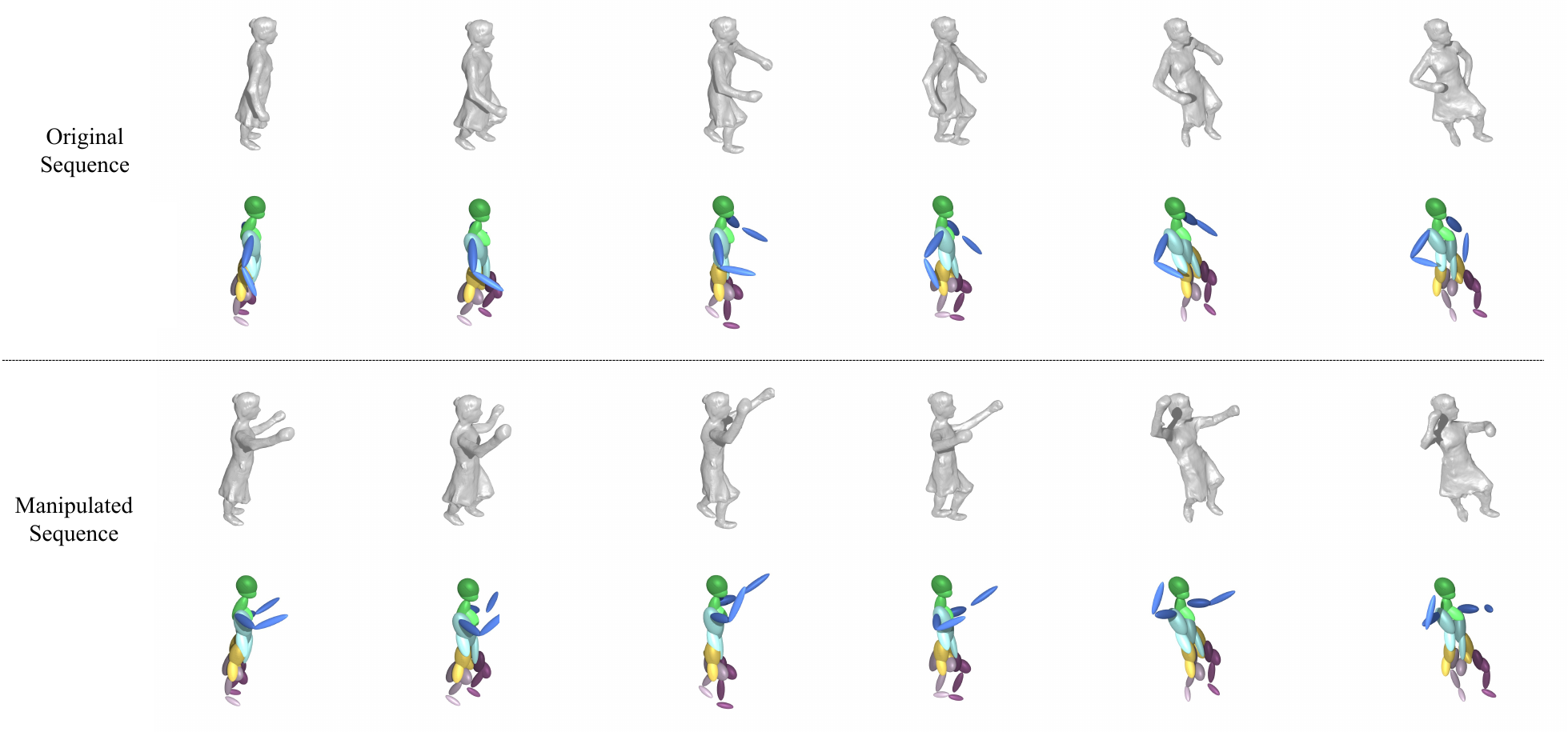}
    \caption{
    Coarse-only manipulation results.
    }
   \label{fig:coarse_manipulation}
\end{figure*}

\textblock{Coarse-only manipulation.}
The decomposition of coarse and fine motions allows coarse-level manipulation of the provided videos while preserving fine-level motions.
\fref{fig:coarse_manipulation} illustrates the results of manipulation.
Specifically, adjusting the parent motions of the arms (colored in blue), which are responsible for controlling both arms, results in the lifting of both arms.
The detailed motions of all child bones are brought from the given sequence, preserving the detailed motions of upper arms, lower arms, and hands.
The decomposition property of our hierarchical deformation model provides an easier and novel way to manipulate 3D models, which is difficult to be achieved in previous approaches.

\begin{figure*}[t]
    \centering
    \includegraphics[width=1\linewidth]{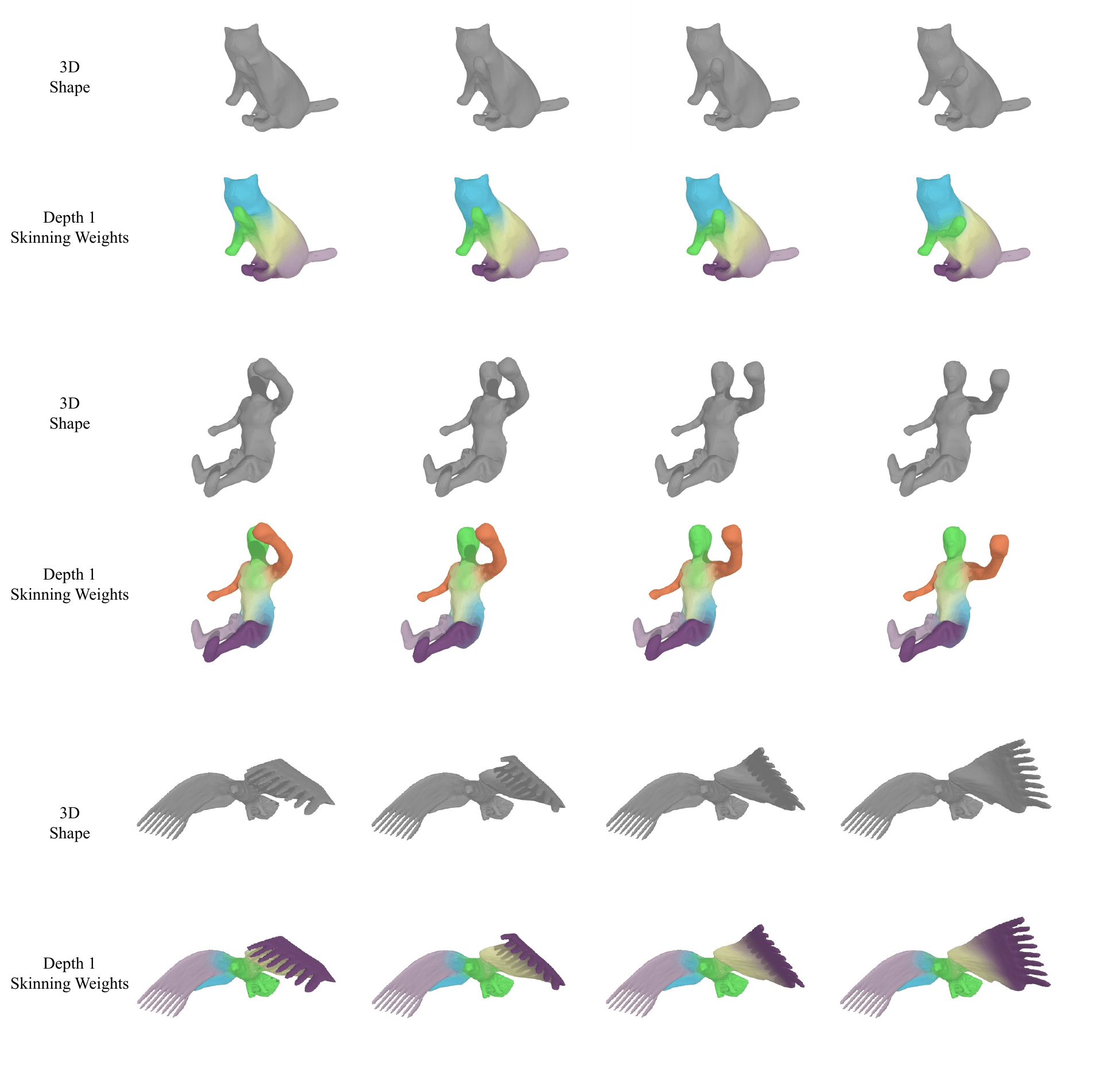}
    \caption{
    Manipulation results on the diverse categories of objects.
    }
   \label{fig:manipulation_bye}
\end{figure*}

\textblock{Manipulation results.}
Lastly, we present manipulation results using both coarse and fine-level manipulations in \fref{fig:manipulation_bye}. 
Such results demonstrate the capability to manipulate 3D models in detail and showcase the ability of our framework to create 3D models with novel poses.
For video results, please refer to the supplementary video.

\section{Societal Impact}
\label{sec:societal}
Our framework presents a range of societal impacts, both positive and negative. 
Positively, it revolutionizes 3D modeling by leveraging casually captured videos, democratizing access to these tools and empowering individuals and small businesses to produce animatable models. 
Additionally, its simplification of the modeling process enhances accessibility, particularly for users with limited technical skills or resources.
However, there are notable concerns regarding potential job displacement, particularly within industries heavily reliant on traditional 3D modeling techniques, as automation may reduce demand for skilled modelers. 
Furthermore, the use of casually captured videos raises privacy concerns, with unauthorized utilization posing risks such as identity theft. 
Additionally, the ease of manipulation facilitated by our framework may exacerbate issues of digital manipulation and misinformation, potentially leading to the spread of false representations and harmful societal consequences.

\clearpage

\end{document}